\documentclass[a4paper]{frontiersSCNS} % for Science, Engineering and Humanities and Social Sciences articles

\usepackage{url,hyperref,lineno,microtype,subcaption}
\usepackage[onehalfspacing]{setspace}
\usepackage{textcomp}

\usepackage{amssymb,amsmath,amsfonts,eurosym,geometry,ulem,graphicx,caption,color,setspace,comment,footmisc,caption,pdflscape,array,hyperref,multirow,url,mathtools,booktabs}

\usepackage{natbib}
\usepackage{apalike}
\usepackage{lscape}
\usepackage{rotating}

\graphicspath{{./pic/}}

\def\keyFont{\fontsize{8}{11}\helveticabold }
\def\firstAuthorLast{Oh {et~al.}} %use et al only if is more than 1 author
\def\Authors{Yee-Hui Oh\,$^{1}$, John See\,$^{2,*}$, Anh Cat Le Ngo\,$^{3}$, Raphael C.-W. Phan\,$^{1,4}$ and Vishnu Monn Baskaran\,$^{5}$} 
\def\Address{
$^{1}$Multimedia University, Faculty of Engineering, Cyberjaya, 63100 Selangor, Malaysia\\
$^{2}$Multimedia University, Faculty of Computing and Informatics, Cyberjaya, 63100 Selangor, Malaysia\\
$^{3}$University of Nottingham, School of Psychology, University Park, Nottingham NG7 2RD, United Kingdom\\
$^{4}$Multimedia University, Research Institute for Digital Security, Cyberjaya, 63100 Selangor, Malaysia\\
$^{5}$Monash University Malaysia, School of Information Technology, Sunway, 47500, Selangor, Malaysia
}

\def\corrAuthor{John See}
\def\corrEmail{johnsee@mmu.edu.my}

\begin{document}

\onecolumn
\firstpage{1}

\title[A Survey of Automatic Facial ME Analysis]{A Survey of Automatic Facial Micro-expression Analysis:\\Databases, Methods and Challenges} 

\author[\firstAuthorLast ]{\Authors} 

%This field will be automatically populated
\address{} %This field will be automatically populated
\correspondance{} %This field will be automatically populated

\extraAuth{}% If there are more than 1 corresponding author, comment this line and uncomment the next one.
%\extraAuth{John See \\ johnsee@mmu.edu.my}

\maketitle

\begin{abstract}
Over the last few years, automatic facial micro-expression analysis has garnered increasing attention from experts across different disciplines because of its potential applications in various fields such as clinical diagnosis, forensic investigation and security systems. Advances in computer algorithms and video acquisition technology have rendered machine analysis of facial micro-expressions possible today, in contrast to decades ago when it was primarily the domain of psychiatrists where analysis was largely manual. Indeed, although the study of facial micro-expressions is a well-established field in psychology, it is still relatively new from the computational perspective with many interesting problems. In this survey, we present a comprehensive review of state-of-the-art databases and methods for micro-expressions spotting and recognition. Individual stages involved in the automation of these tasks are also described and reviewed at length. In addition, we also deliberate on the challenges and future directions in this growing field of automatic facial micro-expression analysis.
\tiny
 \keyFont{ \section{Keywords:} facial micro-expressions, subtle emotions, survey, spotting, recognition, micro-expression databases, spontaneous} 

%All article types: you may provide up to 8 keywords; at least 5 are mandatory.
\end{abstract}

\section{Introduction} \label{sec:introduction}

In 1969,~\cite{ekman1969nonverbal} spotted a quick full-face emotional expression in a filmed interview which revealed a strong negative feeling a psychiatric patient was trying to hide from her psychiatrist in order to convince that she was no longer suicidal. When the interview video was played in slow motion, it was found that the patient was showing a very brief sad face that lasted only for two frames (1/12s) followed by a longer-duration false smile. This type of facial expressions is called micro-expressions (MEs) and they were actually first discovered by~\cite{haggard1966micromomentary} three years before the event happened. In their study, Haggard and Isaacs discovered these “micromomentary” expressions while scanning motion picture films of psychotherapy hours, searching for indications of non-verbal communication between patient and therapist.

MEs are very brief, subtle, and involuntary facial expressions which normally occur when a person either deliberately or unconsciously conceals his or her genuine emotions~\citep{ekman1969nonverbal,ekman2009telling}. Compared to ordinary facial expressions or macro-expressions, MEs usually last for a very short duration which is between 1/25 to 1/5 of a second \citep{ekman2009telling}. Recent research by~\cite{yan2013fast} suggest that the generally accepted upper limit duration of a micro-expression is within 0.5 second. Besides short duration, MEs also have other significant characteristics such as low intensity and fragmental facial action units where only part of the action units of full-stretched facial expressions are presented~\citep{yan2013fast,porter2008reading}. Due to these three characteristics of the MEs, it is difficult for human beings to perceive micro-expressions with the naked eye. 

In spite of these challenges, new psychological studies of MEs and computational methods to spot and recognize MEs have been gaining more attention lately because of its potential applications in many fields, i.e. clinical diagnosis, business negotiation, forensic investigation, and security systems \citep{ekman2009lie,frank2009see,weinberger2010airport}. One of the very first efforts to improve the human ability at recognizing MEs was conducted by Ekman where he developed the Micro-Expression Training Tool (METT) to train people to recognize seven categories of MEs~\citep{ekman2002microexpression}. However, it was found in \citep{frank2009behavior} that the performance of detecting MEs by undergraduate students only reached at most 40\% with the help of METT while unaided U.S. coast guards performed not more than 50 \% at best. Thus, an automatic ME recognition system is in great need in order to help detect MEs such as those exhibited in lies and dangerous behaviors, especially with the modern advancements in computational power and parallel multi-core functionalities. These have enabled researchers to perform video processing operations that used to be infeasible decades ago, increasing the capability of computer-based understanding of videos in solving different real-life vision problems. Correspondingly, in recent years researchers have moved beyond psychology to using computer vision and video processing techniques to automate the task of recognizing MEs.

Although normal facial expression recognition is now considered a well-established and popular research topic with many good algorithms developed \citep{bettadapura2012face,zeng2009survey,sariyanidi2015automatic} with accuracies exceeding 90\%, in contrast the automatic recognition of MEs from videos is still a relatively new research field with many challenges. One of the challenges faced by this field is spotting the ME of a person accurately from a video sequence. As a ME is subtle and short, spotting of MEs is not an easy task. Furthermore, spotting of MEs becomes harder if the video clip consists of spontaneous facial expressions and unrelated facial movements, i.e. eye-blinking, opening and closing of mouth, etc. On the other hand, other challenges of ME recognition include inadequate features for recognizing MEs due to its low change in intensity and lack of complete, spontaneous and dynamic ME databases. 

In the past few years, there have been some noteworthy advances in the field of automatic ME spotting and recognition. However, there is currently no comprehensive review to chart the emergence of this field and summarize the development of techniques introduced to solve these tasks. In this survey paper, we first discuss the existing ME corpora. In our perspective, automatic ME analysis involves two major tasks, namely, ME spotting and ME recognition. ME spotting focuses on finding the occurrence of MEs in a video sequence while ME recognition involves assigning an emotion class label to an ME sequence. For both tasks, we look into the range of methods that have been proposed and applied to various stages of these tasks. Lastly, we discuss the challenges in ME recognition and suggest some potential future directions.

\section{Micro-expression databases} \label{sec:database}

The prerequisite of developing any automatic ME recognition system is having enough labeled affective data. As ME research in computer vision has only gained attention in the past few years, the number of publicly available spontaneous ME databases is still relatively low. Table~\ref{tab:db} gives the summary of all available ME databases to date, including both posed and spontaneous ME databases. The key difference between posed and spontaneous MEs is in the relevance between expressed facial movement and underlying emotional state. For posed MEs, facial expressions are deliberately shown and irrelevant to the present emotion of senders, therefore not really helpful for the recognition of real subtle emotions. Meanwhile, spontaneous MEs are the unmodulated facial expressions that are congruent with an underlying emotional state~\citep{hess1990differentiating}. Due to the nature of the posed and spontaneous MEs, the techniques for inducing facial expressions (for purpose of constructing a database) are contrasting. For the case of posed MEs, subjects are usually asked to relive an emotional experience (or even watching example videos containing MEs prior to the recording session) and perform the expression as well as possible. However, eliciting spontaneous MEs is more challenging as the subjects have to be involved emotionally. Usually, emotionally evocative video episodes are used to induce the genuine emotional state of subjects, and the subjects have to attempt to suppress their true emotions or risk getting penalized.        

According to~\cite{ekman1969nonverbal,ekman2009lie}, MEs are involuntary which could not be created intentionally. Thus, posed MEs usually do not exhibit the characteristics (i.e., the appearance and timing) of spontaneously occurring MEs~\citep{yan2013fast,porter2008reading}. The early USD-HD~\citep{shreve2011macro} and Polikovsky’s~\citep{polikovsky2009facial} databases consist of posed MEs rather than spontaneous ones; hence they do not present likely scenarios encountered in real life. In addition, the occurrence duration of their micro-expressions (i.e., 2/3 s) exceeds the generally acceptable duration of MEs (i.e., 1/2 s). To have a more ecological validity, research interest then shifted to spontaneous ME databases. Several groups have developed a few spontaneous MEs databases to aid researchers in the development of automatic ME spotting and recognition algorithms. To elicit MEs spontaneously, participants are induced by watching emotional video clips to experience a high arousal, aided by an incentive (or penalty) to motivate the disguise of emotions. However, due to the challenging process of eliciting these spontaneous MEs, the number of samples collected for these ME databases is still limited.   

Table ~\ref{tab:db} summarizes the known ME databases in the literature, which were elicited through both posed and spontaneous means. The YorkDDT~\citep{warren2009detecting} is the smallest and oldest database, with spontaneous MEs that also include other irrelevant head and face movements. 
The Silesian Deception database \citep{radlak2015silesian} was created for the purpose of recognizing deception through facial cues. This database is annotated with eye closures, gaze aversion, and micro-expression, or ``micro-tensions'', a phrase used by the authors to indicate the occurrence of rapid facial muscle contraction as opposed to having an emotion category. This dataset is not commonly used in spotting and recognition literature as it does not involve expressions per se; its inception primarily for the purpose of automatic deception recognition.

The SMIC-sub~\citep{pfister2011recognising} database presents a better set of spontaneous ME samples in terms of frame rate and database size. Nevertheless, it was further extended to the SMIC database~\citep{li2013spontaneous} with the inclusion of more ME samples and multiple recordings using different cameras types: high speed (HS), normal visual (VIS) and near-infrared (NIS). However, the SMIC-sub and SMIC databases do not provide Action Unit (AU) (i.e., facial components that are defined by FACS to taxonomize facial expressions) labels and the emotion classes were only based on participants' self-reports. Sample frames from SMIC are shown in Figure \ref{fig:smic_sur}.

The CASME dataset~\citep{yan2013casme} provides a more comprehensive spontaneous ME database with a larger amount of MEs as compared to SMIC. However, some videos are extremely short, i.e. less than 0.2 second, hence poses some difficulty for ME spotting. Besides, CASME samples were captured only at 60 \textit{fps}. An improved version of it, known as CASME II was established to address these inadequacies. The CASME II database~\citep{yan2014casme} is the largest and most widely used database to date (247 videos, sample frames in Figure \ref{fig:casme_hap}) with samples recorded using high frame-rate cameras (200 \textit{fps}). 

To facilitate the development of algorithms for ME spotting, extended versions of SMIC (SMIC-E-HS, SMIC-E-VIS, SMIC-E-NIR), CAS(ME)\textsuperscript{2}~\citep{qu2017cas} and SAMM~\citep{davison2016samm} databases were developed. In SMIC-E databases, long video clips that contain some additional non-micro frames before and after the labeled micro frames were included as well. %SMIC-E has also three parts according to their counterparts in SMIC, which are SMIC-E-HS, SMIC-E-VIS and SMIC-E-NIR.
The CAS(ME)\textsuperscript{2} database (with samples given in Figure \ref{fig:casme2_disgust}) is separated into two parts: Part A contains both spontaneous macro-expressions and MEs in long videos; and Part B includes cropped expression samples with frame from onset to offset. However, CAS(ME)\textsuperscript{2} is recorded using a low frame-rate (25 \textit{fps}) camera due to the need to capture both macro- and micro-expressions. 

In the SAMM database (with samples shown in Figure \ref{fig:samm}), all micro-movements are treated objectively, without inferring the emotional context after each experimental stimulus. Emotion classes are then labeled by trained experts later. In addition, about 200 neutral frames are included before and after the occurrence of the micro-movement, which makes spotting feasible. The SAMM is arguably the most culturally diverse database among all of them. In short, the SMIC, CASME II, CAS(ME)\textsuperscript{2} and SAMM are considered the state-of-the-art databases for ME spotting and recognition that should be widely adopted for research.
%the only few databases which are suitable for ME spotting and recognition tasks and thus they are widely used as the test bed in majority of the research work of MEs. 
%Thus in this survey paper, only the techniques that are applied in these databases are discussed.

The need for data acquired from more unconstrained "in-the-wild" situations have compelled further efforts to provide more naturalistic high-stake scenarios. The MEVIEW dataset \citep{meview} was constructed by collecting mostly poker game videos downloaded from YouTube with a close-up of the player's face. Poker games are highly competitive with players often try to conceal or fake their true emotions, which facilitates likely occurrences of MEs. %Occlusions by hands and cards, and varying ambient illumination are some challenging obstacles. 
With the camera view switching often, the entire shot with a single face in video (averaging 3s in duration) was taken. An METT-trained annotator labeled the onset and offset frames of the ME with FACS coding and emotion types. A total of 31 videos with 16 individuals were collected.

\section{Spotting of Facial Micro-expressions}
\label{sec:spotting}
% As an important non-verbal human behavioural event, facial micro-expressions can be potentially applied in varies domains, which include  business negotiations~\cite{ekman2009telling}, healthcare~\cite{hopf1992localization,cohn2009detecting}, security and behavioural analysis~\cite{weinberger2010airport,frank2009behavior,martin2009philosophy, o2009police,frank2009see}. 
% MEs are rapid, involuntary facial expressions which usually appear in high-stakes environment. Due to the short elapsed time and low intensity, analyzing spontaneous MEs is very challenging. 

Automatic ME analysis involves two tasks: ME spotting and ME recognition. Facial ME spotting refers to the problem of automatically detecting the temporal interval of a micro-movement in a sequence of video frames; and ME recognition is the classification task to identify the ME involved in the video samples. In a complete facial ME recognition system, accurately and precisely identifying frames containing facial micro-movements (which contribute to facial MEs) in a video is a prerequisite for high-level facial analysis (i.e., facial ME recognition). Thus, the automatic facial expression spotting frameworks are developed to automatically search the temporal dynamics of MEs in streaming videos. Temporal dynamics refer to the motions of facial MEs that involve onset(start), apex(peak), offset(end) and neutral phases. Figure \ref{fig:spotting} shows a sample sequence depicting these phases. According to the work by~\cite{valstar2012fully}, the onset phase is the moment where muscles are contracting and appearance of facial changes grows stronger; the apex phase is the moment where the expression peaks (the most obvious); %and there are no more changes in facial appearance due to this particular facial actions; 
and the offset phase is the instance where the muscles are relaxing and the face returns to its neutral appearance (little or no activation of facial muscles). Typically a facial motion shifts through the sequence of neutral-onset-apex-offset-neutral, but other combinations such as multiple apices are also possible.  
% This is usually pre-processed by the coders or annotators manually and resulting only single facial ME sequence in each data. In real time domain, there probably exist several MEs in a streaming video, 

In general, a facial ME spotting framework consists of a few stages: the pre-processing, feature description, and lastly the detection of the facial micro-expressions. The details of each of the stages will be further discussed in the following sections.

\subsection{Pre-processing}
In facial ME spotting, the general pre-processing steps include facial landmark detection, facial landmark tracking, face registration, face masking and face region retrieval. Table.~\ref{tab:tab1} shows a summary of existing pre-processing techniques that are applied in facial ME spotting.

\subsubsection{Facial Landmark Detection and Tracking}
\label{sec:landmark-detection}

Facial landmark detection is the first most important step in the spotting framework to locate the facial points on the facial images. In the field of MEs, two ways of locating the facial points are applied: the manual method and automatic facial landmark detection method. In an early work on facial micro-movement spotting~\citep{polikovsky2009facial}, facial landmarks are manually selected only at the first frame, and fixed in the consecutive frames as they assumed that the examined frontal faces are located relatively in the same location. In their later work~\citep{polikovsky2013facial}, a tracking algorithm is applied to track the facial points that had been manually detected at the first frame throughout the whole sequence. To prevent the hassle of manually detecting the facial points, majority of the recent works ~\citep{davison2015micro,liong2015automatic,wang2016main,xia2016spontaneous,liong2016less,liong2016automatic,davison2016samm,davison2016objective} opt to apply automatic facial landmark detection. Instead of running the detection for the whole sequence of facial images, the facial points are only detected at the first frame and fixed in the consecutive frames with the assumption that these points will only change minimally due to the subtleness of MEs.    

To the best of our knowledge, the facial landmark detection techniques that are commonly employed for facial ME spotting are promoted Active Shape Model (ASM)~\citep{milborrow2014active}, Discriminative Response Maps Fitting (DRMF)~\citep{asthana2013robust}, Subspace Constrained Mean-Shifts (SCMS)~\citep{saragih2009face}, Face++ automatic facial point detector~\citep{facepp} and Constraint Local Model (CLM)~\citep{cristinacce2006feature}. In fact, the promoted ASM, DRMF and CLM are the notable examples of part based facial deformable models. Facial deformable models can be roughly separated into two main categories: holistic (generative) models and part based (discriminative) models. The former applies holistic texture-based facial representation for the generic face fitting scenario; and the latter uses the local image patches around the landmark points for the face fitting scenario. Although the holistic-based approaches are able to achieve impressive registration quality, these representations unfaithfully locate facial landmarks in unseen images, when target individuals are not included in the training set. As a result, part based models which circumvent several drawbacks of holistic-based methods, are more frequently employed in locating facial landmarks in recent years~\citep{asthana2013robust}. 
The promoted ASM, DRMF and CLM are from part based deformable models, however their mechanisms are different. The ASM applies shape constraints and searches locally for each feature point's best location; whereas DRMF learns the variation in appearance on a set of template regions surrounding individual features and updates the shape model accordingly; as for CLM, it learns a model of shape and texture variation from a template (similar to active appearance models), but the texture is sampled in patches around individual feature points. In short, the DRMF is computationally lighter than its counterparts.
%In term of computational capability, DRMF would be faster than the promoted ASM and CLM. This is because the underlying framework of DRMF -- a discriminative regression framework is proposed for learning robust functions from response maps (likelihood maps) to the shape parameters update~\citep{asthana2013robust} which would take shorter time than maximizing the probability of a reconstructed shape that best fit to the test image by adjusting the prior shape model (i.e, the approach applied in ASM). 

Part based approaches mainly rely on optimization strategies to approximate the responses map through simple parametric representations. However, some ambiguities still result due to the landmark's small support region and imperfect detectors. In order to address these ambiguities, SCMS which employs Kernel Density Estimator (KDE) to form a non-parametric representation of response maps was proposed. To maximize over the KDE, the mean-shift algorithm was applied. Despite the progress in automatic facial landmark detection, these approaches are still not considerably robust towards ``in-the-wild" scenarios, where large out-of-plane tilting and occlusion might exist. The Face++ automatic facial point detector was developed by \cite{facepp} to address such challenges. It employs a coarse-to-fine pipeline with neural network and sequential regression, and it claims to be robust against influences such as partial occlusions and improper head pose up to 90$^{\circ}$ tilt angle. The efficacy of the method~\citep{zhang2014coarse} has been tested on the 300-W dataset~\citep{sagonas2013300} (which focuses on facial landmark detection in real-world facial images captured in-the-wild), yielding the highest accuracy among the several recent state-of-the-arts including DRMF. 

In ME spotting research, very few works applied tracking to the landmark points. This could be due to the sufficiency of landmark detection algorithms used (since MEs movements are very minute) or that general assumptions have been made to fix the location of the detected landmarks points. The two tracking algorithms that were reportedly used in a few facial ME spotting works \citep{polikovsky2013facial,moilanen2014spotting,li2017towards} are Auxiliary Particle Filtering (APF)~\citep{pitt1999filtering} and Kanade-Lucas-Tomasi (KTL) algorithm~\citep{tomasi1991detection}. %APF is a filtering method based on Monte Carlo and recursive Bayesian estimation with the introducing auxiliary variable to circumvent the tailed observation densities~\citep{rao2013visual} that occurs in the sequential importance resampling (SIR) algorithm. The advantages of this method is that it can deal successfully with noise, occlusion and clutter, however it is computationally slower~\citep{rao2013visual}. KLT is a gradient-based tracker. It makes use of spatial intensity information to direct the search for the position that yields the best match. It is fast for examining fewer potential matches between the images, however it is not robust to the case of variations in appearance such as illumination and pose object changes, as the KLT may drift to a wrong region of cluttered background under such circumstances~\citep{bagherpour2012upper}.   

\subsubsection{Face Registration}
Image registration is the process of aligning two images -- the reference and sensed images, geometrically. In the facial ME spotting pipeline, registration techniques are applied onto the faces to remove large head translations and rotations that might affect the spotting task. Generally, registration techniques can be separated into two major categories: area-based and feature-based approaches. In each of the approaches, either global mapping functions or local mapping functions are applied to transform the the sensed image to be as close as the reference image.

For area-based (a.k.a. template matching or correlation-like) methods, windows of predefined size or even entire images are utilized for the correspondence estimation during the registration. This approach bypasses the need for landmark points, albeit some restriction to only shift and small rotations between the images~\citep{zitova2003image}. 
%Area-based methods are preferably applied when the images have not prominent details and distinctive information. 
In the work by~\cite{davison2016objective}, a 2D-Discrete Fourier Transform (2D-DFT) was used to achieve face registration. This method calculates the cross-correlation of the sensed and reference images before finding the peak, which in turn is used to find the translation between the sensed and reference images. Then, the process of warping to a new image is performed by piece-wise affine (PWA) warping.
%This method has two-folds benefits~\citep{zitova2003image}. Firstly, it is robust against the correlated and frequency dependent noise and non-uniform, time varying illumination disturbances. And secondly, the computation time savings are more significant if the images that are to be registered are large. But, there is a probability of mismatching the details of the sensed and reference images due to its non-saliency. 
% Compared to global mapping model such as affine transform and non-reflective similarity transform as used in the literature~\cite{davison2015micro, davison2016samm,wang2016main}, local mapping model such as piecewise affine transform can handle deformations locally.

For feature-based approach to face registration, salient structures which include region features, line features and point features are exploited to find the pairwise correspondence between the sensed and reference images. Thus, feature-based approach are usually applied when the local structures are more significant than the information carried by the image intensities. In some ME works \citep{shreve2011macro,moilanen2014spotting,li2017towards}, the centroid of the two detected eyes are selected as the distinctive point (also called control points) and exploited for face registration by using affine transform or non-reflective similarity transform. The consequence of such simplicity entails their inability to handle deformations locally. A number of works \citep{xu2016microexpression,li2017towards} employed Local Weighted Mean (LWM) \citep{lwm} which seeks to find a 2-D transformation matrix using 68 facial landmark points of a model face (typically from the first frame). In another work by~\cite{xia2016spontaneous}, Procrustes analysis is applied to align the detected landmark points in frames. It determines a linear transformation (such as translation, reflection, orthogonal rotation, and scaling) of the points in sensed images to best conform them to points in the reference image. Procrustes analysis has several advantages: low complexity for easy implementation and it is practical for similar object alignment~\citep{ross2004procrustes}. However, it requires a one-to-one landmark correspondence and the convergence of means is not guaranteed. 

Instead of using mapping functions to map the sensed images to the reference images, a few studies~\citep{moilanen2014spotting,shreve2011macro,li2017towards} correct the mis-alignment by rotating the faces according to the angle between the pair of lines that join the centroids of the two detected eyes to the horizontal line. In this mechanism, errors can creep in if the face contours of the sensed and reference face images are not consistent with one another, or that the subject's face is not entirely symmetrical to begin with.  

% To address this problem,~\cite{davison2015micro, davison2016samm} applied affine transform; and~\cite{wang2016main} used non-reflective similarity transform to map the sensed face images to the reference face image. Both the affine transform and non-reflective similarity transform belong to global mapping models, in fact non-reflective similarity transform is a subset of affine transform. For non-reflective similarity transform, it may include rotation, scaling, and translation; whereas for affine transform, it has extra functions such as shearing and reflection.

Due to the diversity of face images with various types of degradations to be registered, it is challenging to fix a standard method that is applicable to all conditions. Thus, the choice of registration method should correspond to the assumed geometric deformation of the sensed face image.

\subsubsection{Masking}
%Masking is also known as spatial filtering. In general, the filtering operation is performed directly on the image to either for blurring, noise reduction or increase the sharpness of the image. 
In the facial ME spotting task, a masking step can be applied onto the face images to remove noise caused by undesired facial motions that might affect the performance of the spotting task. In the work by~\cite{shreve2011macro}, a static mask ('T'-shaped) was applied on the face images to remove the middle part of the face that includes the eyes, nose and mouth regions. Eye regions were removed to avoid the noise caused by eye cascades and blinking (which is not considered a facial micro-expression); the nose region is masked as it is typically rigid, which might not reveal much significant information even with it; and mouth region is excluded since opening and closing of the mouth introduces undesired large motion. It is arguable if too much meaningful information may have been removed from the face area in the masking steps introduced in~\citep{shreve2011macro, shreve2014automatic}, as the two most expressive facial parts (in the context of MEs) are actually located near the corner of the eyebrow and mouth areas. Hence, some control is required to prevent excluding too much meaningful information. 
%around the corners of the eyes and mouth, the landmark points for the specific parts are used in the masking process. 
Typically, specific landmark points around these two areas are used as reference or boundary points in the masking process.

In the work by~\cite{liong2016automatic}, the eye regions are masked to reduce false spotting of the apex frame from long videos. They observed that eye blinking motion is significantly more intense than that of micro-expression motion, thus masking is necessary. To overcome potential inaccurate landmark detection, a 15-pixel margin was added to extend the masked region. %With the focus firmly on retrieving only meaningful information for the spotting task,
Meanwhile, ~\cite{davison2016objective} applied a binary mask to obtain 26 FACS-based facial regions that include the eyebrows, forehead, cheeks, corners around eyes, mouth, regions around mouth and etc. The regions are useful for the spotting task as each of these regions contain a single or a group of AUs, which will be triggered when the ME occurs. It is also worth mentioning that a majority of works in the literature still do not include a masking pre-processing step. 
% In the later work~\citep{shreve2014automatic}, the mouth and eyes were masked to remove non-rigid motions.

\subsubsection{Face Region Retrieval}
From psychological findings on concealed emotions~\citep{porter2008reading}, it was revealed that facial micro-expression analysis should be done on the upper and lower halves of the face separately instead of considering the entire face. This finding substantiated an earlier work~\citep{rothwell2006silent}, whereby ME recognition was also performed on the segmented upper and lower parts of the face. \cite{duan2016recognizing} later showed that the eye region is much more salient than the whole face or mouth region for recognizing micro-expressions, in particular happy and disgust expressions. Prior knowledge from these works encourage splitting of the face into important regions for automatic facial micro-expression spotting. 

In the pioneering work of spotting facial MEs~\citep{shreve2009towards}, the face was segmented into 3 regions: the upper part (which includes the forehead), middle part (which includes the nose and cheeks) and the lower part (which include the mouth); and each was analyzed as individual temporal sequences. In their later work~\citep{shreve2011macro}, the face image is further segmented into 8 regions: forehead, left and right of the eye, left and and right of cheek, left and right of mouth and chin. Each of the segments is analyzed separately in sequence. With the more localized segments, tiny changes in certain temporal segments could be observed. However, unrelated edged features such as hair, neck and edge of the face that are present in the localized segments might induce noise and thus affect the extracted features. Instead of splitting the face images into few segments,~\cite{shreve2014automatic} suggested to separate the face images into four quadrants, and each of the quadrant is analyzed individually in the temporal domain. The reason behind this is because of the constraint on locality as facial micro-expressions are restricted to appear in at most two bordering regions (i.e., first and second quadrant, second and third quadrant, third and forth quadrant and the first and fourth quadrant) of the face~\citep{shreve2014automatic}.  

% The main reason is that micro-expressions can only appear in one region of the face due to the low intensity of the elapsed motions of facial micro-expressions~\citep{shreve2009towards,shreve2014automatic}. 

Another popular facial segmentation method is splitting the face into a specific number ($m \times n$) of blocks~\citep{moilanen2014spotting,davison2015micro,wang2016main,davison2016samm,li2017towards}. In the blocking representation, the motion changes in each block could by observed and analysis independently. However, with the increasing in the number of blocks (i.e., $m \times n$), the computation load increases accordingly. Besides, features such as hairs and edges of face that appear in the blocks will affect the final feature vectors as these elements are not related to the facial motions. 

%in addition, considering the contribution of each block equally is not logical as the information contained in each block varies 
   
A unique approach to facial segmentation for ME spotting is to split the face by Delaunay triangulation~\citep{davison2016objective}. It gives more freedom to the shape that defines the regions of the face. %With this method, it can reduce the unrelated regions such as hair and neck. 
Unfortunately, areas of the face that are not useful for ME analysis such as the cheek area may still be captured within the triangular regions. To address this problem, more recent methods partition the face into a few region-of-interests (ROIs)~\citep{polikovsky2009facial,polikovsky2013facial,liong2015automatic,liong2016automatic, liong2016less,davison2016objective}. The ROIs are regions that correspond to one or more FACS action units (AUs). As such, these regions contain rigid facial motions when the muscles (AUs) are activated. Some studies~\citep{liong2015automatic,liong2016automatic, liong2016less,davison2016objective} show that ROIs are more effective compared to the use of the entire face in constraining the salient locations for spotting.
% (i.e., involving of unrelated regions such as hair, neck and edges of face)

%\subsection{Feature Descriptor}
\subsection{Facial Micro-expression Spotting}
Facial micro-expression spotting, or \textit{``micro-movement''} spotting (a term coined by \cite{davison2016samm}) refers to the problem of automatically detecting the temporal interval of a micro-movement in a sequence of video frames. Current approaches for spotting facial micro-movement can be broadly categorized into two groups: classifier-based methods (supervised / unsupervised) and rule-based (use of thresholds or heuristics) methods. There are many possible dichotomies; this survey discusses some early ideas, followed by two distinct groups of works -- one on spotting ME movement or window of occurrence, another on spotting the ME apex. %Classifier-based methods employ either a supervised classifier (such as SVM) or an unsupervised approach (such as $k$-means clustering); whereas rule-based methods refer to threshold techniques where a threshold is used as the indicator to spot the facial micro-movement. 
A summary of the existing techniques for spotting facial micro-expressions (or micro-movements) are depicted in Table.~\ref{tab:spottingTable}.

\subsubsection{Early works}

In the early works by~\cite{polikovsky2009facial,polikovsky2013facial}, 3D-HOG was adopted to extract the features from each of the regions in the ME videos. Then, $k$-means clustering was used to cluster the features to particular AUs within predefined facial cubes. ``Spotting'' was approached as a classification task: each frame is classified to neutral, onset, apex or offset, and compared with ground truth labels. The classification rates achieved were satisfactory, in the range of 68\%--80\%. %for onset, apex and offset were 78.13\% (80.02\% with transition tags), 68.34\% (70.99\% with transition tags) and 79.84\% (81.85\% with transition tags) respectively. 
%The transition tags are designed to correspond to boundary frames for which phase they are ambiguous to be classified to. 
Although their method could potentially contribute to facial micro-movement spotting by locating the four standard phases described by FACS, there are two glaring drawbacks. First, their method was only tested on posed facial ME videos, which are not a good representation of spontaneous (naturally induced) facial MEs. Secondly, the experiment was run as a classification task in which the frames were clustered into one of the four phases; this is highly unsuitable for real-time spotting. 
The work of~\cite{wu2011} also treats the spotting task as a classification process. Their work uses Gabor filters and the GentleSVM classifier to evaluate the frames. From the resulting label of each frame, the duration of facial micro-expressions were measured according to the transition points and the video frame-rate. Subsequently, they are only considered as ME when their durations last for 1/25$s$--1/5$s$. They achieved very high spotting performance on the METT training database~\citep{ekman2003micro}. However, this was not convincing on two counts; first, only 48 videos were used in the experiments, and second, the videos were synthesized by inserting a flash of micro-expression in the middle of a sequence of neutral face images. In real-world conditions, frame transitions would be much more dynamic compared to the abrupt changes that were artificially added.

%~\cite{xia2016spontaneous} applied the random walk model to compute the probability of frames having micro-expressions by considering the geometrical deformation correlation between frames in a temporal window. In their work, they assumed the frame predicted negative as positive if there are certain number of frame predicted as positive in the temporal window. The performance of the proposed method heavily relies on the accuracy of landmark spotting, which make it not feasible in real-life scenarios where the head movements and ambient light conditions could degrade the landmark detection. 

Instead of treating the spotting task as frame-by-frame classification, the works of ~\cite{shreve2009towards,shreve2011macro} are the first to consider the temporal relation from frame-to-frame and employ a threshold technique to locate spontaneous facial MEs. This follows a more objective method that does not require machine learning. Their works are also the first in the literature to attempt spotting both macro- (i.e., ordinary facial expressions) and micro-expressions from videos. In their work, optical strain, which represents the amount of deformation incurred during motion, was computed from selected facial regions. Then, the facial MEs are spotted by tracking the strain magnitudes across frames following these heuristics: %and used in plotting the strain magnitude occurring across frame. In the plot, the spotted sequences could only be considered as facial MEs if 
(1) strain magnitude exceeds the threshold (calculated from the mean of each video) and is significantly larger than that of the surrounding frames, and (2) the duration of the detected peak can only last at most 1/5th of a second. %The threshold values were calculated from the mean strain magnitude of each videos. 
A 74\% true positive rate and 44\% false positive rate was achieved in the spotting task. However, a portion of data used in their experiments were posed, while some of them (Canal-9 and Found Videos databases) were not published or are currently defunct. In their later work~\citep{shreve2014automatic}, a peak detector was applied to locate sequences containing MEs based on strain maps. However, the details of the peak detector and threshold value were not disclosed. 

\subsubsection{Movement spotting}

Micro-expression movements can be located by identifying a "window" of occurrence, typically marked by a starting or \textit{onset} frame, and an ending or \textit{offset} frame. In the work by~\cite{moilanen2014spotting}, the facial motion changes were modeled by feature difference (FD) analysis of appearance-based features (i.e., LBP) that incorporates the Chi-Square ($\chi^2$) distance to form the FD magnitudes. Only the top $1/3$ of total blocks (per frame) with the greatest FD values were chosen and averaged to an initial feature value representing the frame. The contrasting difference vector is then computed to find relevant peaks from across the sequence.
%local magnitude variation and background noise. 
Spotted peak frames (i.e., the peaks that exceed the threshold) are compared with the provided ground truth frames; and considered true positive if they fall within the span of $k/2$ frames (where $k$ is half of the interval frames in the window) before the onset and after the offset. The proposed technique was tested on CASME-A, CASME-B and SMIC-VIS-E, achieving a true positive rate of 52\%, 66\% and 71\%, respectively. 

The same spotting approach was adopted by~\cite{li2017towards} and tested on various spontaneous facial ME databases: CASME II, SMIC-E-HS, SMIC-E-VIS and SMIC-E-NIR. %This is the first report of spotting on existing spontaneous ME datasets.
This work also indicated that LBP consistently outperforms HOOF in all the datasets with higher AUC (area-under-the-ROC-curve) values and lower false positive rates. To spot facial micro-expressions on the new CAS(ME)\textsuperscript{2} database, the same spotting approach~\citep{moilanen2014spotting} is adopted by~\cite{wang2016main}. Using their proposed main directional optical flow (MDMD) approach, ME spotting performance on the CAS(ME)\textsuperscript{2} is 0.32, 0.35 and 0.33 for recall, precision and F1-score, respectively. For all these works~\citep{moilanen2014spotting,wang2016main,qu2017cas,li2017towards}, the threshold value for peak detection is set by taking the difference between the mean and max value of the contrasting difference vector and multiplying it by a fraction in the range of [0,1]. Finally, this value is added with the mean value of the contrasting difference vector to denote the threshold. By these calculations, at least one peak will always be detected as the threshold value will never exceed the maximum value of the contrasting difference vector. This could potentially result in misclassification of non-ME movements since it will \textit{always} detect a peak. Besides, pre-defining the ME window intervals (which obtains the FD values) may not augur well with videos captured at different frame rates. 
%could make the approach not flexible as each data might need different window size. 
To address the potentiality of a false peak, these works~\citep{moilanen2014spotting,wang2016main,qu2017cas,li2017towards,davison2015micro} proposed to compute the baseline threshold based on a neutral video sequence from each individual subject in the datasets. %Therefore, the FD vector is supposed not to exceed the baseline threshold if there is no facial movement. 

In the work of \cite{davison2015micro}, all detected sequences which are less than 100 frames are denoted as true positives, in which eye blinks and eye gaze are included; while peaks that are detected but not coded as a movement are classed to false positives. The approach achieved scores of 0.84, 0.70 and 0.76 for recall, precision and F1-measure, respectively on the SAMM database. In their later works,~\cite{davison2016samm,davison2016objective} introduced ``individualised baselines'', which are computed by taking a neutral video sequence for the participants and using the $\chi^2$ distance to get an initial feature for the baseline sequence. The maximum value of this baseline feature is identified as the threshold. This improved their previous attempt by a good margin. 

A number of innovative approaches were proposed. \cite{patel2015spatiotemporal} computed optical flow vectors over small local regions and integrated them into spatiotemporal regions to find the onset and offset times. In another approach, ~\cite{xia2016spontaneous} applied random walk model to compute the probability of frames containing MEs by considering the geometrical deformation correlation between frames in a temporal window. \cite{duque2018micro} designed a system that is able to differentiate between MEs and eye movements by analyzing the phase variations between frames based on the Riesz Pyramid.

\subsubsection{Apex spotting}

Besides spotting facial micro-movements, a few other works focused on spotting a specific type of ME phase, particularly the \textit{apex} frame~\citep{liong2015automatic,liong2016automatic,liong2016less,yan2017measuring}. 
%that focus on apex frame spotting for facial ME videos.
The apex frame, which is the instant indicating the most expressive emotional state in an ME sequence, is believed to be able to effectively reveal the true expression for the particular video. In the work by~\cite{yan2017measuring}, the frame that has the largest feature magnitude was selected as the apex frame. A few interesting findings were revealed: CLM (which provides geometric features) is especially sensitive to contour-based changes such as eyebrow movement, and LBP (which produces appearance features) is more suitable for detecting changes in appearance such as pressing of lips; however, OF is the most all-rounded feature as it is able to spot the apex based on the resultant direction and movement of facial motions. A binary search method was proposed by~\cite{liong2015automatic} to automatically locate the apex frame in a video sequence. %Generally in the conventional method, the apex frame can be spotted by locating the maximum peak (as applied in~\citep{yan2017measuring}). 
%The authors found that the maximum peak from a video sequence may not necessarily be an apex frame, due to inaccurate feature values. 
By observing that the apex frames are more likely to appear in areas concentrated with peaks, %from the per-frame analysis on CASME II dataset.
%With this,  
the proposed binary search method iteratively partitions the sequence into two halves, by selecting the half that contains a higher sum of feature difference values. 
%The half which produces lower magnitude is eliminated and the search continues on the remaining half 
This is repeated until a single peak is left. The proposed method reported a mean absolute error (MAE) of 13.55 frames and standard error (SE) of 0.79 on CASME II using LBP difference features. A recent work by \cite{ma2017region} used Region HOOF (RHOOF) based on 5 regions of interests (ROIs) for apex detection, which resulted in more robust results.
  
\subsection{Performance Metrics}

The ME spotting task is akin to a binary detection task (ME is present / not present), hence typical performance metrics can be used. \cite{moilanen2014spotting} encouraged the use of a Receiver Operating
Characteristic (ROC) curve, which was adopted in most subsequent works \citep{patel2015spatiotemporal,xia2016spontaneous,li2017towards}. In essence, the spotted peaks, which are obtained based on a threshold level, will be compared against ground truth labels to determine whether they are true or false spots. If one spotted peak is located within the frame range of [onset - $\frac{N-1}{4}$, offset + $\frac{N-1}{4}$] of a labeled ME clip, the spotted sequence ($N$ frames centered at the peak) will be considered as a true positive ME; otherwise the $N$ frames of spotted sequence will be counted as false positive frames. The specified range considers a tolerance interval of 0.5 seconds, which corresponds to the presumed maximum duration of MEs. To obtain the ROC curve, true positive rate (TPR) and false positive rate (FPR) are computed as follows:
\begin{equation}
	\text{TPR} = \frac{\text{Number of frames of correctly spotted MEs}}{\text{Total number of ground truth ME frames from all samples}}
\label{eq:tpr}
\end{equation}
\begin{equation}
	\text{FPR} = \frac{\text{Number of incorrectly spotted frames}}{\text{Total number of non-ME frames from all samples}}
\label{eq:fpr}
\end{equation}

Recently, \cite{sliding2017} proposed a micro-expression spotting benchmark (MESB) to standardize the performance evaluation of the spotting task. Using a sliding window based multi-scale evaluation and a series of protocols, they recognize the need for a fairer and more comprehensive method of assessment. Taking a leaf out of object detection, the Intersection over Union (IoU) of the detection set and ground truth set was proposed to determine if a sampled sub-sequence window is positive or negative for ME (threshold set at 0.5). 

Several works that focused on the spotting of the apex frame \citep{yan2014quantifying,liong2015automatic,liong2016automatic,liong2016less} used Mean Absolute Error (MAE) to compute how close are the estimated apex frames to the ground-truth apex frames:
\begin{equation}
	\text{MAE} = \frac{1}{N}\sum_{i=1}^{N}|e_{i}|
\label{eq:mae}
\end{equation}
When spotting is performed on the raw long videos, \cite{liong2016automatic} introduced another measure called Apex Spotting Rate (ASR), which calculates the success rate in spotting apex frames within the given onset and offset range of a long video. An apex frame is scored 1 if it is located between the onset and offset frames, and 0 otherwise:
\begin{flalign}
	\text{ASR} &= \frac{1}{N}\sum_{i=1}^{N}\delta_{i} \\
    \text{where} \quad \delta & = \begin{cases} 
      1, & \text{if} f^*\in(f_{i,\text{onset}},f_{i,\text{offset}}) \nonumber \\
      0, & \text{otherwise} 
   \end{cases}
\label{eq:asr}
\end{flalign}

\section{Facial Micro-expression Recognition} \label{sec:recognition}
ME recognition is a task that classifies an ME video into one of the universal emotion classes (e.g., Happiness, Sadness, Surprise, Anger, Contempt, Fear and Disgust). However, due to difficulties in the elicitation of micro-expressions, not all classes are available in the existing datasets. Typically, the emotion classes of the collected samples are unevenly distributed; some are easier to elicit hence they have more samples collected. 

Technically, a recognition task involves feature extraction and classification. However, a pre-processing stage could be involved prior to the feature extraction to enhance the availability of descriptive information to be captured by descriptors. In this section, all the aforementioned steps are discussed.   

% Emotions expressed by the human face can be categorized into basic and non-basic. Basic emotions refer to the affect model developed by Ekman and his colleagues \cite{ekman1971constants} where this model has seven classes: happiness, sadness, surprise, fear, anger disgust and contempt. These basic emotions are emotions that have been scientifically proven by many psychologists to have a certain facial expression associated with it and their facial expressions are universal across different cultures in the world \cite{ekman1971constants}. Additionally, non-basic emotions refer to secondary and tertiary emotions, such as relief, embarrassment and guilty, which are unlikely to observe through facial expression \cite{parrott2001emotions}. Since MEs reveal genuine emotions of a person, correct classification of MEs would be a crucial task for automatic ME recognition system. The few steps involved in an automatic ME recognition system are pre-processing, feature extraction and classification.

% The main challenge of this task is solving the problem of adequate facial features in micro-expressions for emotion classification as these expressions are too rapid and subtle. No machine learning algorithms can perform well with inadequate features. Lately, literature on the classification of facial expressions from well segmented videos that are already divided into temporal sequences containing the micro-expressions has been increasing. The fundamental components of the micro-expression recognition system consist of pre-processing, facial feature extraction, and classification. 

\subsection{Pre-processing}

A number of fundamental pre-processes such as face landmark detection and tracking, face registration and face region retrieval, have all been discussed in Section~\ref{sec:spotting} for the spotting task. Most recognition works employ similar techniques as those used for spotting, i.e. ASM \citep{milborrow2014active}, DRMF \citep{asthana2013robust}, Face++ \citep{facepp} for landmark detection; LWM \citep{lwm} for face registration. Meanwhile, division of the facial area into regions is a step often found within various feature representation techniques (discussed in Section \ref{sec:representations}) to further localize features that change subtly. Aside from these known pre-processes, two essential pre-processing techniques have been instrumental in conditioning ME data for the purpose of recognition. We discuss these two steps which involve \textit{magnification} and \textit{interpolation} of ME data.

%that aim to discover or reveal much more descriptive information from the native face images are explored as well to overcome some challenges in micro-expression recognition. 

The uniqueness of facial micro-expressions is in its subtleness, which is one of reasons why recognizing them automatically is very challenging. As the intensity levels of facial ME movements are very low, it is extremely difficult to discriminate ME types among themselves. One solution to this problem is to exaggerate or magnify these facial micro-movements. In recent works \citep{park2015subtle,zarezadeh2016micro,wang2016effective,li2017towards}, the Eulerian Motion Magnification (EMM)~\citep{wu2012eulerian} method was employed to magnify the subtle motions in the ME videos. The EMM method extracts the frequency bands of interest from the different spatial frequency bands obtained from the decomposition of an input video, by using band-pass filters; these extracted bandpass signals at different spatial level are amplified by a magnification factor $\alpha$ to magnify the motions.~\cite{li2017towards} demonstrated that the EMM method helps to enlarge the difference between different categories of micro-expressions (i.e, inter-class difference); thus the recognition rate is increased. However, larger amplification factors may cause undesirable amplified noise (i.e. motions that are not induced by MEs), which may degrade recognition performance. To prevent over-magnifying ME samples,~\cite{le2016eulerian} theoretically estimated the upper bounds of effective magnification factors. Besides, the authors also compared the performance of the amplitude-based Eulerian motion magnification (A-EMM) and phase-based Eulerian motion magnification (P-EMM); with the %technique was carried out to identify which state-of-the-art technique works better in boosting the micro-expression recognition rate. 
%Experiments showed that 
A-EMM performing marginally better than a rather noise-ridden P-EMM. %which is rather susceptible to noise. 
To deal with the distinctive temporal characteristic of different ME classes, a magnification scheme was proposed by \cite{park2015subtle} to adaptively select the most discriminative frequency band needed for EMM to magnify subtle facial motions. 
%based on the temporal characteristic of the MEs. 
%The proposed scheme was reportedly able to improve the effectiveness of motion magnification for the ME recognition task. 
%Subsequent works later \citep{li2017towards} also reported significant increases when using Eulerian motion magnification as a pre-processing step. 
A recent work by \cite{le2018micro} showed that Global Lagrangian Motion Magnification (GLMM) can contribute towards better recognition capability compared to local Eulerian based approaches, particularly at higher magnification factors.
%Overall Eulerian motion magnification techniques help to enhance the performance of ME recognition a significant rate.

Another concern for ME recognition is with the uneven length (or duration) of ME video samples. In fact, it can contribute to two contrasting scenarios: (a) the case of short duration videos, which restricts the application of the feature extraction techniques which require varied temporal window size (e.g., LBP-based methods that can form binary patterns from varied radius); and (b) the case of long duration videos, whereby redundant or replicated frames (due to high frame rate capture) could deteriorate the recognition performance. To solve the problem, the temporal interpolation method (TIM) is applied to either up-sample (clips that are too short) or down-sample (clips that are too long) clips to produce clips of similar frame lengths. 

Briefly, TIM takes original frames as input data to construct a manifold of facial expressions; then it samples on the manifolds for a particular number of output frames (refer to \cite{zhou2011towards} for detailed explanation). It is shown by \cite{li2017towards} that modifying the frame length of ME videos can improve the recognition performance if the number of interpolated frames are small. However, when the interpolated frames are increased, the recognition performance is somewhat hampered due to over-interpolation. Therefore, the appropriate interpolation of the ME sequence is vital in preparation for recognition. 
%Compared to TIM which only temporally and regularly down-samples image sequences, 
An alternative technique Sparsity-Promoting Dynamic Mode Decomposition (DMDSP)~\citep{jovanovic2014sparsity} was adopted by~\cite{le2015subtle,le2016sparsity} to select only significant dynamics in MEs to form sparse structures. From the comprehensive experimental results shown in \citep{le2016sparsity}, DMDSP achieved better recognition performance compared to TIM (on similar features and classifiers) due to its ability to keep only the significant temporal structures while eliminating irrelevant facial dynamics. 

While the aforementioned pre-processing techniques showed positive results in improving ME recognition, yet these methods will notably lengthen the computation time of the overall recognition process. For a real-time system to be feasible, this cost has to be taken into consideration.

% Le Ngo et al. \cite{le2015subtle} discovered that micro-expressions recorded at high frame rate, e.g. 200 fps, create a lot of redundant frames with no significant facial motion while inducing rapid illumination changes and noise to the video. They later eliminated this redundancy by using Sparsity-Promoting Dynamic Mode Decomposition (DMDSP) \cite{jovanovic2014sparsity} and the output was a dynamically condensed sequence, containing only the significant frames of micro-expressions. DMDSP is the variant of Dynamic Mode Decomposition (DMD) \cite{schmid2010dynamic} which is a data processing technique that is capable of extracting coherent structures with a single temporal frequency or dynamic mode from a numerical data sequence. On the other hand, DMDSP finds the least number of DMD modes that have the most influence on approximating the original data sequence. A clearer explanation of DMDSP and its algorithm can be found in \cite{jovanovic2014sparsity}. The authors did a comprehensive study on DMDSP method and they reported that performance of the micro-expression recognition system improved significantly using the synthesized video sequence given by DMDSP compared to the video sequence given by TIM \cite{le2016sparsity}.

\subsection{Facial Micro-expression Representations}
\label{sec:representations}
In the past few years, research in automatic ME analysis have been much focused on the problem of ME recognition: given an ME video sequence/clip, the purpose of recognition is to estimate its emotion label (or class). Table~\ref{tab:recogTable} summarizes the existing ME methods in the literature. From the perspective of feature representations, they can be roughly divided into two main categories: \textit{single-level} approaches and \textit{multi-level} approaches. Single-level approaches refer to frameworks that directly extract feature representations from the video sequences; while for multi-layer approaches, the image sequences are first transformed into another domain or subspace prior to feature representation to exploit other kinds of information to describe MEs.  

Feature representation is a transformation of raw input data to a succinct form; typically in face processing, representations can be from two distinct categories: geometric-based or appearance-based \citep{zeng2009survey}.
%that can be effectively exploited in machine learning tasks. 
Specifically, geometric-based features describe the face geometry such as the shapes and locations of facial landmarks; whereas appearance-based features describe intensity and textural information such as wrinkles, furrows and other patterns that are caused by emotion. However from previous studies in facial expression recognition~\citep{zeng2009survey,fasel2003automatic}, it is observed that appearance-based features are better than geometric-based features in coping with illumination changes and mis-alignment error. Geometric-based features might not be as stable as appearance-based features as they need precise landmark detection and alignment procedures. For these similar reasons, appearance-based feature representations have become more popular in the literature on ME recognition %because of their robustness against generic image processing issues such as illumination variation and registration errors as well as their implementation simplicity. 

\subsubsection{LBP-based methods}

Among appearance-based feature extraction methods, local binary pattern on three orthogonal planes (LBP-TOP) is widely applied in many works~\citep{yan2014casme,li2013spontaneous,guo2014micro,le2014spontaneous,le2015subtle,le2016sparsity,zheng2016relaxed,le2016eulerian,wang2016effective,adegun2016automatic}. Most existing datasets (SMIC, CASME II, SAMM) have all reported the LBP-TOP as their baseline evaluation method. LBP-TOP is an extension of its low-level representation, local binary pattern (LBP)~\citep{ojala2002multiresolution}, which describes local texture variation along a circular region with binary codes which are then encoded into a histogram. LBP-TOP extracts features from local spatio-temporal neighbourhoods over three planes: the spatial (XY) plane similarly to the regular LBP, the vertical spatio-temporal (YT) plane and the horizontal spatio-temporal (XT) plane; this enables LBP-TOP to dynamically encode temporal variations. %However, it is noted that there is a gap to achieve a high-performance ME recognition by using LBP-TOP. 

Subsequently, several variants of LBP-TOP were proposed for the ME recognition task.~\cite{wang2014lbp} derived {Local Binary Pattern - Six Interception Points (LBP-SIP) from LBP-TOP by considering only the 6 unique points lying on three intersecting lines of the three orthogonal planes as neighbor points for constructing the binary patterns. By reducing redundant information from LBP-TOP, LBP-SIP reported better performance than LBP-TOP in this task. A more compact variant, LBP-MOP~\citep{wang2015efficient} was constructed by concatenating the LBP features from only three mean images, which are the temporal pooling result of the image stacks along the three orthogonal planes. The performance of LBP-MOP was comparable to LBP-SIP, but with its computation time dramatically reduced. While LBP considers only pixel intensities, spatio-temporal completed local quantized patterns (STCLQP)~\citep{huang2016spontaneous} exploited more information containing sign, magnitude and orientation components. To address the sparseness problem (in most LBP variants), specific codebooks were designed to reduce the number of possible codes to achieve better compactness. 

Recent works have yielded some interesting advances. \cite{huang2017spontaneous} proposed a new binary pattern variant called spatio-temporal local Radon binary pattern (STRBP) that uses Radon transform to obtain robust shape features. \cite{ben2017learning} proposed an alternative binary descriptor called Hot Wheel Patterns (HWP) (and its spatio-temporal extension HWP-TOP) to encode the discriminative features of both macro- and micro-expressions images. A coupled metric learning algorithm is then used to model the shared features between micro- and macro-expression information.

\subsubsection{Optical flow-based methods}

As suggested in several studies (e.g. \citep{li2017towards}), the temporal dynamics that reside along the video sequences are found to be essential in improving the performance of ME recognition. As such, optical flow (OF)~\citep{horn1981determining} based techniques, which measure the spatio-temporal changes in intensity, came into contention as well.
%able to measure the spatio-temporal changes of intensity to locate the matching pixel in the next frame 

In the work by~\cite{xu2016microexpression}, a proposal to extract only principal directions of the OF maps was purportedly to eliminate abnormal OF vectors that resulted from noise or illumination changes. 
%This was based on the assumption that facial facial motion moves in roughly the same spatial and temporal directions when the observation window is small enough. 
A similar concept of exploiting OF in the main direction was employed by~\cite{liu2016main} to design main directional mean optical flow (MDMO) features. MDMO is a ROI-based OF feature, which considers both local statistic (i.e., the mean of OF vectors in the bin with the maximum count in each ROI) and its spatial location (i.e., the ROI to which it belongs). %By applying normalization, the feature dimension of MDMO is fixed to just $72$ regardless of the number of frames in an image sequence. 
Unlike the aforementioned works which exploited only the single dominant direction of OF in each facial region,~\cite{allaertconsistent} determined the consistent facial motion, which could be in multiple directions from a single facial region. The assumption was made based on the fact that facial motions spread progressively due to skin elasticity, hence only the directions that are coherent in the neighboring facial regions are extracted to construct a consistent OF map representation. 

Motivated by the use of optical strain (OS) for ME spotting \citep{shreve2009towards,shreve2014automatic}, \cite{liong2014optical} proposed to leverage on its strengths for ME recognition. OS is derived from OF by computing the normal and shear strain tensor components of the OF. This enables the capture of small and subtle facial deformation. 
%To leverage the strengths of OS to ME recognition,~\cite{liong2014optical} proposed the use of OS as the feature representation. 
In their work, the OS magnitude images are temporally pooled to form a single pooled OS map; then the resulting map is max-normalized and resized to a fixed smaller resolution before transforming into a feature vector that represent the video. To emphasize the importance of active regions, the authors ~\citep{liong2014subtle} proposed to weight local LBP-TOP features with different weights which were generated from the temporal mean-pooled OS map. This allows regions that actively exhibit MEs to be given more significance, hence increasing the discrimination between emotion types.%In their later work~\citep{liong2016spontaneous}, the row-vectorized OS map features were concatenated with the LBP-TOP with OS weights as the final features. %However, the performance is not as good as the previous individual works. This might due to the limitation of the proposed features. 
In a more recent attempt, ~\cite{liong2016less} proposed a Bi-Weighted Oriented Optical Flow (BI-WOOF) descriptor which applies two schemes to weight the HOOF descriptor locally and globally. Locally, the magnitude components were used to weight the orientation bins within each ROI; the resultant locally weighted histograms are then weighted again (globally) by multiplying with the mean optical strain (OS) magnitude of each ROI. Intuitively, a larger change in the pixel's movement or deformation will contribute towards a more discriminative histogram. Instead of considering the whole image sequences, the authors also demonstrated promising recognition performance using only two frames (i.e., the onset frame and the apex frame) instead of using whole sequences. This was able to reduce the processing time by a large margin.%As such, the processing time can be further reduced and make the system running on real-time. 

\cite{zhang2017micro} proposed to aggregate the histogram of the oriented optical flow (HOOF)~\citep{chaudhry2009histograms} with LBP-TOP features region-by-region to generate local statistical features. In their work, they revealed that fusing of local features within each ROI can capture more detailed and representative information than globally done. In the work by~\cite{happy2017fuzzy}, fuzzy histogram of optical flow orientation (FHOFO) was proposed for ME recognition. In HFOFO, the histograms are only the collection of orientations without being weighted by the optical flow magnitudes; the assumption was made that MEs are so subtle that the induced magnitudes should be ignored. They also introduced a fuzzification process that considers the contribution of an orientation angle to its surrounding bins based on fuzzy membership functions; as such smooth histograms for motion vector are created. 

%%%%%%%%%%%%%
\subsubsection{Other methods}

Aside from methods based on low-level features, there are also numerous techniques proposed to extract other types of feature representations.~\cite{lu2014delaunay} proposed a Delaunay-based temporal coding model (DTCM) to encode the local temporal variation (in grayscale values) in each subregion obtained by Delaunay triangulation and preserve the ones with high saliency as features. %In the work, the saliency was computed by the personal Standard deviation (STD) in which the magnitude of STD reflects the strength of facial movement in the corresponding facial subregion. By employing the proposed saliency analysis method, subregions with higher saliency, which are assumed to be highly related to the MEs are selected; on the other hand subregions with lower saliency are assumed to be irrelevant to MEs are discarded. 
In the work of~\cite{li2017towards}, the histogram of image gradient orientation (HIGO), which is a degenerate variant of HOG, was employed in the recognition task. It uses simple vote rather than weighted vote when counting the responses of the gradient orientations. As such, it could depress the influence of illumination contrast by ignoring the magnitude. The use of color space was also experimented in the work of~\cite{wang2015micro}, where LBP-TOP features were extracted from Tensor Independent Color Space (TICS). In TICS, the three color components (R, G and B) were transformed into three uncorrelated components which are as independent as possible to avoid redundancy and thus increase the recognition performance. The Sparse Tensor Canonical Correlation Analysis (STCCA) representation proposed by \cite{wang2016sparse} offers a solution to mitigate the sparsity of spatial and temporal information in a ME sequence.

Signal components such as magnitude, phase and orientation can be exploited as features for ME recognition. \cite{oh2015monogenic} proposed a monogenic Riesz wavelet framework, where the decomposed magnitude, phase and orientation components (which represent energy, structural and geometric information respectively) are concatenated to describe MEs. In their extended work~\citep{oh2016intrinsic}, higher-order Riesz transform was adopted to exploit the intrinsic two-dimensional (i2D) local structures such as corners, junctions and other complex contours. They demonstrated that i2D structures are better representative parts than i1D structures (i.e., simple structures such as lines and straight edges) in describing MEs. By supplementing the robust Bi-WOOF descriptor \citep{liong2016less} with Riesz monogenic phase information derived from the onset-apex difference image \citep{liong2017micro}, recognition performance can be further boosted.

Integral projections are an easy way of simplifying spatial data to obtain shape information along different directions. The LBP-Integral Projection (IP) technique proposed by~\cite{huang2015facial} applies the LBP operator on these projections. A difference image is first computed from successive frames (to remove face identity) before it is projected into two parts: vertical projection and horizontal projection. This method was found to be more effective than directly using features derived from the original appearance information. 
In their extended work~\citep{xiaohua2017discriminative}, original pixel information is replaced by extracted subtle emotion information as input for generating spatio-temporal local binary pattern with revisited integral projection (STLBP-RIP) features. To further enhance the discriminative power of these features, only features with the smallest Laplacian scores are selected as the final feature representation. 

% need to specially discuss about the introducing of RPCA and Feature selection to address the problems faced.
A few works increase the significance of features by means of excluding irrelevant information such as pose and subject identity, which may obstruct salient emotion information. Robust principal component analysis (RPCA)~\citep{wright2009robust} was adopted in~\citep{wang2014micro,huang2016spontaneous} to extract subtle emotion information for feature extraction. 
In~\citep{wang2014micro}, the extracted subtle emotion information was encoded by local spatio-temporal directional (LSTD) to extract more detailed spatio-temporal directional changes on the \textit{x}, \textit{y} and \textit{t} directions from each plane (XY, XT and YT). 
%To remove the correlation of coefficients from these six directions, these coefficients were decorrelated to from decorrelated LSTD (DLSTD) features. 
\cite{lee2017multimodal} proposed an interesting use of Multimodal Discriminant Analysis (MMDA) to orthogonally decompose a sample into three modes or "identity traits" (emotion, gender and race) in a simultaneous manner. Only the essential emotion components are magnified before the samples are synthesized and reconstructed. 

Recently, numerous new works have begun exploring other forms of representation and mechanisms. \cite{he2017multi} proposed a strategy to extract low-level features from small regions (or cubes) of a video by learning a set of class-specific feature mappings.
%To obtain descriptive and discriminative features, in the work by~\cite{he2017multi} a feature learning framework is proposed to learn the low-level features (e.g., LBP-TOP, LBP-MOP and OSW) from each of the small regions of a video. Several class-specific feature mappings are learned from the low-level features under the multi-task learning mechanism, where each of the class-specific feature mappings pulls the features of the same class together and pushes the features from other classes farther. To improve the discriminative power of the learning features, two contrastive weighting schemes are proposed: (a) features obtained from the active facial regions are assigned with higher weights; and (b) features obtained from the in-active regions are assigned with higher weight. By increasing the weightage on in-active regions, the subtle discrepancies between different MEs in in-active regions were emphasized.
~\cite{jia2017macro} devised a macro-to-micro transformation model based on singular value decomposition (SVD) to recognize MEs by utilizing macro-expressions as part of the training data.
This overcomes the lack of labeled data in MEs databases.  %In the proposed model, singular value decomposition (SVD) was employed to achieve the macro-to-micro transformation. To selectively pick up the important and efficient patches (e.g., the regions around of eyes, lips and cheek muscles) that are significant for distinguishing different kinds of emotions, the Group LASSO~\citep{yuan2006model} was employed. 
%Motivated by the desirable characteristics of Gabor wavelets which show spatial locality and orientation selectivity, ~\cite{zheng2017micro} proposed to employ 2D Gabor filter to extract features for ME recognition. To deal with the subtlety of MEs, 
There were various recent attempts at casting the recognition task as one arising from a different problem. \cite{zheng2017micro} formulated it as a sparse approximation problem and presented the 2D Gabor filter and sparse representation (2DSGR) technique for feature extraction. \cite{zhu2018coupled} drew inspiration from similarities between MEs and speech to propose a transfer learning method that projects both domain signals to a common subspace. In a radical move, ~\cite{davison2017objective} proposed to re-group MEs based on Action Units (AUs) instead of by emotion categories, which are arguably susceptible to bias in self-reports used during the construction of dataset. %This is the best way of recognizing MEs as it removes the potential bias of human reporting in constructing dataset. 
Their experimental results on CASME II and SAMM showed that recognition performance should be higher than what is currently expected from other works that used emotion labels. 
%that the proposed objective classes (i.e., By organizing the Aus of the into different classes) outperformed the original class (i.e., emotion labels) in both CASME II and SAMM datasets in all feature extraction techniques(i.e., LBP-TOP, HOOF and HOG3D).  

\subsection{Classification}
% Most affect recognition systems rely on generic classification models such as SVM. Affect recognition has its own specific dynamics and recent studies aimed at tailoring statistical models for affect recognition. The new models address several issues such as modeling the temporal variations of emotions or expressions, personalizing existing models, modeling statistical dependencies between expressions or utilizing domain knowledge by exploiting correlations among affect dimensions.

The last stage in an ME recognition task involves the classification of the emotion type. Various types of classifiers have been used for the task of ME recognition such as $k$-Nearest Neighbor ($k$-NN), Support Vector Machine (SVM), Random Forest (RF), Sparse Representation Classifier (SRC), Relaxed K-SVD, Group Sparse Learning (GSL) and Extreme Learning Machine (ELM).
%Generally, classifiers can be broadly separated into two categories: supervised classifiers and unsupervised classifiers. Supervised classifiers employ supervised learning where an algorithm is used to learn the mapping function from the input to the output; and unsupervised classifiers model the underlying structure or distribution in the data in order to learn more about the data. 
From the literature, the most widely used classifier %we can tell that SVM is the most popular classifier in the ME recognition.
%The majority of practical classifiers are supervised. The notable example of the supervised classifiers that has been widely employed in classifying Facial MEs 
is the SVM. SVMs are computational algorithms that construct a hyperplane or a set of hyperplanes in a high or infinite dimensional space~\citep{cortes1995support}. During the training of SVM, the margins between the borders of different classes are sought to be maximal. Compared to other classifiers, SVMs are robust, accurate and very effective even in cases where the number of training samples is small. On the contrary, two other notable classifiers -- RF and $k$-NN are seldom used in the ME recognition task. Although the RF is generally quicker than SVM, it is prone to overfit when dealing with noisy data. The $k$-NN uses an instance-based learning process which may not be suitable for sparse high-dimensional data such as face data.
%RF is an ensemble of unpruned trees created by using bootstrap samples of the training data and random feature selection in tree induction. Prediction is made by aggregating (majority vote or averaging) the predictions of the ensemble. %Compared to SVM, the runtime of RF is faster and they are able to deal with unbalanced and missing data. However, RF is overfitting-prone especially when dealing with noisy data. Among all the classifiers, k-NN can be considered the simplest machine learning classifier. It is a type of instance-based learning where the function is only approximated locally and all computation is deferred until classification. It is effective when dealing with large training data; however, the distance based learning is not clear as which type of distance to use to produce the best results.

To deal with the sparseness of MEs, several works tried using relaxed K-SVD, SRC and GSL techniques for classification. However, each of these methods tackle the sparseness of MEs differently. The relaxed K-SVD \citep{zheng2016relaxed} learns a sparse dictionary to distinguish different MEs by minimizing the variance of sparse coefficients. %In Relaxed K-SVD~\citep{zheng2016relaxed}, the reconstruction error and the classification error are taken into consideration, while the variance of sparse coefficients is minimized to address the similarity of same classes and distinctiveness of different classes in sparse coefficients. 
The SRC used in \cite{zheng2017micro} represents a given test sample as a sparse linear combination of all training samples; hence the sparse nonzero representation coefficients are likely to concentrate on training samples that are of the same class as the test sample. %In SRC, the truncation residuals are used as the classification criterion. The SRC judges the test samples to \textit{i}th class if the \textit{i}th residual is the smallest.
% new work
A Kernelized GSL \citep{zong2018learning} is proposed to facilitate the process of learning a set of importance weights from hierarchical spatiotemporal descriptors that can aid the selection of the important blocks from various facial blocks.
Neural networks can offer a one-shot process (feature extraction and classification), with a remarkable ability to extract complex patterns from data. % to derive meaning from complicated or imprecise data, neural network can be used to extract patterns and detect trends that are too complex to be noticed by other machine learning techniques. 
However, a substantial amount of labeled data is required to properly train a neutral network without overfitting it, resulting in it being less favorable for ME recognition since labeled data is limited. The ELM~\citep{huang2006extreme}, which is naturally just feed-forward network with a single hidden layer was used by \cite{adegun2016automatic} to classify MEs. %and a promising recognition was achieved. 
%From the literature, we can tell that SVM is the most popular classifier in the ME recognition. However, there has been no agreement on which classifier is the most suitable for ME classification.  

\subsection{Experimental Protocol \& Performance Metrics}

The original dataset papers \citep{yan2014casme,li2013spontaneous,davison2016samm} all propose the adoption of the Leave-One-Subject-Out (also known as ``LOSO'') cross-validation as the default experimental protocol. This is done with consideration that the samples were collected by eliciting the emotions from a number of different participants (i.e., $S$ number of subjects). As such, cross validation should be carried out by withholding a particular subject $s$ while the other $S-1$ subjects are used in the training step. This removes the potential identity bias that may arise during the learning process; a subject that is being evaluated could have been seen and learned in the training step. A number of other works used the Leave-One-Video-Out (``LOVO") cross-validation protocol instead, which exhaustively divides all samples into $S$ number of possible train-test partitions. This protocol is deemed to avoid irregular partitioning but is often likely to overestimate the performance of the classifier. A few works opted to report their results using their own choice of evaluation protocol, such as an evenly distributed sets \citep{zhang2017micro}, random sampling of test partition \citep{jia2017macro} and 5-fold cross validation \citep{adegun2016automatic}.
Generally, the works in literature can be categorized into these three groups, as shown in Table \ref{tab:recogTable}.

The ME recognition task reports the typical performance metric of \textit{Accuracy}, which is commonly used in other image/video recognition problems. A majority of works in the literature report the Accuracy metric, which is simply the number of correctly classified video sequences over the total number of video sequences in the dataset. However, due to the imbalanced nature of the ME datasets which was first discussed by \cite{le2014spontaneous}, Accuracy scores can be highly skewed towards classes that are larger as classifiers tend to learn poorly from classes that are less represented. Consequently, it makes more sense to report the \textit{F1-Score} (or F-measure), which is the harmonic mean of the \textit{Precision} and \textit{Recall}:

\vspace{-1em}
\begin{align}
	\centering
	\text{\textit{F1-Score}} &= 2 \cdot \frac{Precision \cdot Recall}{Precision + Recall}\\
    \text{\textit{Precision}} &= \frac{tp}{tp+fp}\\
    \text{\textit{Recall}} &= \frac{tp}{tp+fn}
	\label{eq:metrics}
\end{align}
where $tp$,  $fp$ and  $fn$ are the number of true positives, false positives, false negatives, respectively. The overall performance of a method can be reported by \textit{macro-averaging} across all classes (i.e. compute scores for each class, then average them), or by \textit{micro-averaging} across all classes (i.e. summing up the individual $tp$, $fp$ and $fn$ in the entire set before computing scores).

%taking into consideration the \textit{Precision} and \textit{Recall} scores of all classes separately, and then micro-averaging it to obtain the final scores. 

\section{Challenges} \label{sec:discussion}
The studies reviewed in Sections~\ref{sec:database},~\ref{sec:spotting} and~\ref{sec:recognition} show the progress in the research work in ME analysis. However, there is still considerable room for improvement in the performance of ME spotting and recognition. In this section, some recognized problems in existing databases and challenging issues in both tasks are discussed in detail. 

\subsection{Micro-expression Databases}
Acquiring valuable spontaneous ME data and their ground truth is far from being solved. Among the various affective states, certain emotions (such as happiness) are relatively easier to be elicited compared to others (e.g., fear, sadness, anger)~\citep{coan2007handbook}. Consequently, there is an imbalanced distribution of samples per emotion and number of samples per subject. This could be biased towards particular emotions that constitute a larger portion of the training set. To address this issue, a more effective way of eliciting affective MEs (especially to those are relatively difficult) should be discovered. Social psychology has suggested creative strategies for inducing affective expressions that are difficult to elicit~\citep{coan2007handbook}. %This could be a good reference to design proper and suitable elicitation methods for MEs.
Some works have underlined the possibility of using other complementary information from the body region \citep{song2013learning} or instantaneous heart rate from skin variations \citep{gupta2018exploring} to better analyze micro-expressions.

Almost all the existing datasets contain a majority of subjects from one particular country or ethnicity. Though we know the fact that basic facial expression are universal across the cultural background, nevertheless subjects from different backgrounds may express differently towards the same elicitation, or at least with different intensity level as they may have different ways of expressing an emotion. Thus, a well-established database should comprise a diverse range of ethnic groups to provide better generalization for experiments.

Although much effort has been paid towards the collection of databases of spontaneous MEs, some databases (e.g. SMIC) lack important metadata such as FACS. It is generally accepted that human facial expression data need to be FACS coded. The main reason being that FACS AUs are objective descriptors and independent of subjective interpretation. Moreover, it is also essential to report the reliability measure of the inter-observers (or inter-coders) involved in the labeling of data. 
%availability of data labeling could incur subsequent issue: the reliability of the coding. Thus, the inter-observer (i.e., inter-coder) reliability should be reported to show how extend the data are accurately labeled.

Considering the implementation of real-life applications of ME recognition in the near future, existing databases that are constructed under studio environments, may not best represent MEs exhibited in real-life situations. Thus, developing and introducing real-world ME databases could bring about a leap of progress in this domain.

\subsection{Micro-expression Spotting}
Recent work on the spotting of MEs have achieved promising results on successfully locating the temporal dynamics of micro-movements; however, there is room for improvement as the problem of spotting MEs remains a challenging task to date. 

\textit{Landmark detection}. Even though the facial landmark detection algorithms have made remarkable progress over the past decade, the available landmark detectors are not always accurate or steady. The unsteadiness of face alignment based on imprecise facial landmarks may result in significant noise (i.e., rigid head movements and eye gaze) associated with dynamic facial expressions. This in turn increases the difficulty in detecting the correct MEs. Thus, a more advanced robust facial landmark detection is required to correctly and precisely locate the landmark points on the face. %As such, the rigid motion such as head movement can be removed after the precise face alignment. 

\textit{Eyes: To keep or not keep?} To avoid the intrusion of eye blinks, majority of works in the literature simply mask out the eye regions. However, according to some findings~\citep{vaidya2014eye,zhao2011facial,lu2015combining,duan2016recognizing}, the eye region is one of the most discriminative regions for affect recognition. As many spontaneous MEs involving muscles around eye regions, there is a need to differentiate between the eye blinks corresponding to certain expressions and those that are merely irrelevant facial motions. In addition, the onsets of the many MEs also temporally overlap with eye blinks~\citep{li2017towards}. Thus, this warrants a more robust approach at dealing with overlapping occurrences of facial motions.  
%masking off the eye regions does not appear to be an ideal approach for improving micro-expression spotting.

\textit{Feature-based or rule-based?} The few studies~\citep{yan2017measuring,liong2015automatic} investigated the effectiveness of individual feature descriptors in capturing the micro-movements for the ME spotting task. They have showed that micro-movements that are induced from different facial components actually resulted in motion changes from different perspectives such as appearance, geometric and etc. For example, lifting up or down the eyebrows results in a clear contour change (geometrical), which could be effectively captured by geometric-based feature descriptors; pressing of lips could cause the variation in appearance but not the position, and thus appearance-based feature descriptors can capture these changes. Interestingly, they reported that motion-based features such as optical flow based features outperformed appearance-based and geometric-based features in the ME spotting. %In the existing work, motion-based features that are employed in the spotting task are generated from the generic of OF. However, OF is estimated based on three assumptions: brightness constancy, spatial coherence and temporal persistence. 
The problem remains that the assumptions made by optical flow methods are likely to be violated in unconstrained environments, rendering real-time implementation challenging.    

Majority of existing efforts toward the spotting of MEs employ rule-based approaches that rely on thresholds. Frames with magnitude exceeding the pre-defined threshold value are the frames (i.e., the temporal dynamics) where ME appears. However, prior knowledge is required to set the appropriate threshold for distinguishing the relevant peaks from local magnitude variation and background noise. This is not really practical in the real-time domain. Instead,~\cite{liong2015automatic} designed a simple divide-and-conquer strategy, which does not require a threshold to locate the temporal dynamics of MEs. Their method finds the apex frame based on a high concentration of peaks. 

\textit{Onset and offset detection}. Further steps should also be considered to locate the onset and offset frames of these ME occurrences. %From the literature, there is no work that establishes the criteria for defining the onset and offset frames for MEs. 
While it is relatively easier to identify the peaks and valleys of facial movements, the onset and offset frames are much more difficult to determine. The task of locating the onset and offset frames will be even tougher when dealing with real-life situations where facial movements are continuously changing.
%and might not depict clear and sharp turning points. 
Thus, the indicators and criteria for determining the onset and offset frames need to be properly defined and further studied. Spotting the ME onset and offset frames is a crucial step which can lead to automatic ME analysis.  

\subsection{Micro-expression Recognition}

In the past few years, much effort has been done towards ME recognition, including developing new features to better describe MEs. However, due to the short elapsed duration and low intensity of MEs, there is still room for improvement towards achieving satisfactory accuracy rates. This could be due to several possible reasons.

\textit{Block selection}. In most works, block-based segmentation of a face to extract local information is a common practice. Existing efforts employed block-based segmentation of a face without considering the contribution from each of the blocks. Ideally, the contribution from all blocks should be varied, whereby the blocks containing the key facial components such as eyebrows, eyes, mouth and cheek should be highlighted as the motion changes at these regions convey meaningful information from differentiating different MEs. Higher weights can be assigned to those regions that contain key facial components to enhance the discriminative power. Alternatively, the discriminative features from the facial blocks can be selected through a learning process; the recent work of \cite{zong2018learning} offers a solution to this issue.  
%while less discriminative features can be discarded to avoid disruption from less related information. 

\textit{Type of features}. Since the emergence of the ME recognition works, many different feature descriptors have been proposed for MEs. Due to the characteristic of the feature descriptors, the extracted features might carry different information (e.g., appearance, geometric, motion, etc). For macro-expressions, it has been shown in~\citep{zeng2009survey,fasel2003automatic} that geometric-based features performed poorer than appearance- and motion-based features as they are highly dependent on the precision of facial landmark points. However, recent ME works~\citep{huang2015facial,xiaohua2017discriminative} show that shape information is arguably more discriminative for identifying certain MEs. Perhaps different features may carry meaningful information for different expression types. This should be carefully exploited and taken into consideration during feature extraction process. %describing the facial motion changes of different MEs discriminatingly.

\textit{Deep learning}. The advancement of Deep Learning has prompted the community to look for new ways of extracting better features. However, a crucial ingredient to this remains as to the feasible amount of data necessary to train a model that does not over-fit easily; the small scale of data (lack of ME samples per category) and the imbalanced distribution of samples are the primary obstacles. Recently an approach by \cite{patel2016selective} made an attempt to utilize deep features transferred from pretrained ImageNet models. The authors deemed that fine-tuning the network to the ME datasets is not plausible (insufficient data) and opted for a feature selection scheme. Some other works \citep{kim2016micro,peng2017dual} have also begun exploring the use of deep neural networks by encoding spatial and temporal features learned from network architectures that are relatively ``shallower'' than those used in the ImageNet challenge \citep{imagenet}. This may be a promising research direction in terms of advancing the features used for this task.

% on cross-database recognition
\textit{Cross-database recognition}. Another on-going development that challenges existing experimental presuppositions is cross-database recognition. This setup mimics a realistic setting where training and test samples may come from different environments. Current recognition performance based on single databases, is expected to plunge under such circumstances. \cite{zong2017learning,zong2018domain} proposed a domain regeneration (DR) framework, which aims to regenerate micro-expression samples from source and target databases. The authors aptly point out that much is still to be done to discover more robust algorithms that work well across varying domains. The first ever Micro-Expression Grand Challenge \citep{yap2018facial} was held with special attention given to the importance of cross-database recognition settings. Two protocols -- Hold-out Database Evaluation (HDE) and Composite Database Evaluation (CDE), were proposed in the challenge, using the CASME II and SAMM databases. The reported performances \citep{merghani2018facial,peng2018macro,khor2018enriched} were poorer than most other works that apply only to single databases, indicating that future methods need to be more robust across domains.

\subsection{Experiment Related Issues}
\textit{Evaluation Protocol}. An important issue that should be addressed in ME recognition is how the data is evaluated. Due to the different evaluation protocols used in existing works, a fair comparison among these works could not be adequately established. Currently, the two popular evaluation protocols that are widely applied in ME recognition are: leave-one-video-out cross-validation (LOVOCV) and leave-one subject-out cross validation (LOSOCV). The common \textit{k}-fold cross-validation is not suitable as the current publicly available spontaneous ME datasets are highly imbalanced \citep{le2014spontaneous}. The number of samples per subject and the number of samples per emotion class in these datasets vary quite considerably. For instance, in the CASME II dataset, the number of samples that belong to the 'Surprise' class is 25 compared to the 102 samples of the 'Others' class; while the difference in the number of samples for 'Subject 08' and 'Subject 17' are 8 and 34 respectively. As such, with \textit{k}-fold cross-validation, the fairness in evaluation is likely to be questionable. The same goes with employing LOVOCV, where only one video sample is left out as the test sample while the remaining samples are used for training; subsequently, the average accuracy across all folds is taken as the final result. This can possibly introduce additional biases on certain subjects that have more representation during the evaluation process. Moreover, the performance of such a protocol typically over-estimates the actual classifier performance due to a substantially large training set. We would stress that the LOSOCV protocol is a more convincing evaluation protocol as it separates the samples of the test set based on the subject identity. As such, the training model is not biased towards the identity of the subject (akin to face recognition instead). Naturally, this protocol also limits the ability of methods to learn the intrinsic micro-expression dynamics of each subject. The intensity and manner of which micro-expressions are shown may differ from person to person, hence compartmentalizing a subject altogether may inhibit the modeling process.

\textit{Performance Metrics}. Besides the usage of evaluation protocol, the choice of performance metrics is also crucial to understanding the actual performance of automatic ME analysis. Currently, two performance metrics are used most widely: the Accuracy rate and F1-score. While the Accuracy rate is straightforward in calculation, it does not give an adequate reflection of the effectiveness of a classifier as it is susceptible to heavily skewed data (uneven distribution of samples per emotion class), a characteristic found in most current datasets. %which is found in almost all existing datasets. 
Also, the Accuracy rate merely shows the average ``hit rate" across all classes; thus the performance of the classifier that deals with each emotion class is not revealed. It is a much preferred practice to report confusion matrices for better understanding of its per-class performances. From thereafter, performance metrics such as F1-score, Precision and Recall provide a better measure of a classifier's performance when dealing with imbalanced datasets~\citep{sokolova2009systematic,le2014spontaneous}. The overall F1-score, Precision and Recall scores should be micro-averaged based on the total number of true positives, false positives and false negatives. 

% In a classification task, precision is the number of true positive samples divided by the number of classified positive samples; recall is the number of true positive samples divided by the number of ground-truth positive samples; and F1-score is the overall indicator that considers both the precision and recall, whereby it reflects relations between classified positive samples and ground-truth positives samples. 

\textit{Emotion class}. There are several existing works considering different number of emotion classes instead of using the emotion classes provided by the databases. For instance, in the works by~\cite{zheng2017micro} and \cite{wang2015micro}, the authors considered only three or four emotion labels (i.e., Positive, Negative, Surprise and/or Others) instead of the original emotion labels of the CASME II (i.e., Happiness, Surprise, Disgust, Repression and Others). Due to the reduction in the number of emotion classes considered, the classification task could be relatively simpler compared to those that have more emotion classes. As a result, higher performances were reported but this also inhibits these works from fair benchmarking against other works on the merit of their methods. It is important to note also that the grouping of classes may be biased toward negative categories since there is only one positive category (Happiness).

Recently, \cite{davison2017objective} challenged the current use of emotion classes by proposing the use of \textit{objective classes}, which are determined by restructuring these new categories around the Action Units (AUs) that have been FACS coded. Samples from the two most recent FACS coded datasets, CASME II and SAMM, were re-grouped into these objective classes for their use. The authors argued that emotion classification requires the context of the situation for an interpreter to make a meaningful interpretation, while relying on self-reports \citep{yan2014casme} can also cause further unpredictability and bias. Although FACS coding can objectively assign AUs to specific muscle movements of the face but the emotion type becomes less obvious. \cite{lim2017fuzzy}, through their fuzzy modeling, provided some insights as to why the emotional content in ME samples are non-mutually exclusive as they may contain traces of more than one emotion type.

\section{Conclusion} \label{sec:conclusion}
Research on the machine analysis of facial MEs has witnessed substantial progress in the last few years as several new spontaneous facial MEs databases were made available to aid automatic analysis of MEs. This has spiked the interest of the affective and visual computing community with a good number of promising methods making headways in both automatic ME spotting and recognition tasks. This necessitates a comprehensive review of recent advances to better taxonomize the increasing number of existing works. In addition, this paper also summarizes the issues that have not received sufficient attention, but are crucial for feasible machine interpretation of MEs. Among the important issues that are yet to be addressed in the field of ME spotting:
\begin{itemize}
  \item Handling macro movements: Differentiating between larger, macro facial movements such as eye blinks and twitches, for better spotting of the onset of MEs,
  \item Developing more precise spotting techniques that can cope with various head poses and camera views: Extension of current constrained environments towards more real-time ``in-the-wild" settings will provide a major leap in practicality. 
  \item Establishing a firm criteria for defining the onset and offset frames for MEs: This allows ME short sequences to be extracted from long videos, which in turn, can be classified into emotion classes.
\end{itemize}

\noindent For the ME recognition task, there are a few issues that deserve the community's attention:
\begin{itemize}
  \item Excluding irrelevant facial information: As MEs are very subtle, it is a great challenge to remove other image perturbations caused by face alignment and slight head rotations which may interfere with the MEs.
  \item Improving feature representations: Encoding subtle movements are difficult even when feature representations are rich, due to limitations in the amount of data that we have. 
  \item Initiating cross-database evaluation: Evaluating within single databases often gives a false impression of a method's performance, especially when existing databases lack diversity.
\end{itemize}

\singlespacing
\setlength\bibsep{0pt}
\bibliographystyle{apa}
\bibliography{ref}

\clearpage
%%%%%%%%%%%%%%%%%%%%%%%%%

% funding information to be provided separately during submission
\section*{Acknowledgement}

This work was supported in part by the Telekom Malaysia funded projects UbeAware, 2beAware and ParaDigm, and MOHE Grant FRGS/1/2016/ICT02/MMU/02/2 Malaysia.

\clearpage
%%%%%%%%%%%%%%%%%%%%%%%%%

\onehalfspacing

\section*{Tables} \label{sec:tab}
\addcontentsline{toc}{section}{Tables}

\begin{table}[htbp]
  \centering
  \caption{Micro-expression Databases}
   \resizebox{\textwidth}{!}{%
   %\scalebox{0.70}{
    \begin{tabular}{l|c|c|c|c|c|c|c|c|c}
    \midrule
    \textbf{Databases} & \textbf{Subset} & \textbf{Subjects} &  \textbf{Samples} &  \textbf{Frames} &  \textbf{Type$^\ast$} &  \textbf{FACS} &  \textbf{Emotion} & \textbf{Expression} & \textbf{Frame} \\
  & & & & \textbf{per sec} & & \textbf{Coded} & \textbf{Classes} & & \textbf{Annotations} \\ \hline
    \midrule
    USF-HD &       & -     & 100   & 30    & P & No    & 6 & macro/micro & - \\ \hline
    Polikovsky's &       & 10    & 42    & 200   & P & No    & 6 & micro & -  \\ \hline
    YorkDDT &       & 9     & 18    & 25    & S & No    & 2 & micro & - \\ \hline
    Silesian Deception$^\dagger$ &       & 101     & 101    & 100    & S & No    & - & macro/micro & eye closures,\\ 
    & & & & & & & & & gaze aversion,\\
    & & & & & & & & & micro-tensions \\ \hline
    SMIC-sub &       & 6     & 77    & 100   & S & No    & 3 & micro & - \\ \hline
    \multirow{6}[0]{*}{SMIC } & HS    & 16    & 164   & 100   & S & No    & 3 & \multirow{3}{*}{micro} & \multirow{3}{*}{-} \\
          & VIS   & 8     & 71    & 25    & S & No    & 3 & \\
          & NIR   & 8      & 71    & 25    & S & No    & 3  & \\ \cline{2-10} 
          & E-HS  & 16    & 157   & 100   & S & No    & 3 & \multirow{3}{*}{micro} & \multirow{3}{*}{onset,offset} \\
          & E-VIS & 8     & 71    & 25    & S & No    & 3  & \\
          & E-NIR & 8     & 71    & 25    & S & No    & 3  & \\ \hline
    CASME  &       & 19    & 195   & 60    & S & Yes   & 7 & micro & onset,offset,apex \\ \hline
    CASME II &       & 26    & 247   & 200   & S & Yes   & 5 & micro & onset,offset,apex \\ \hline
    \multirow{2}[0]{*}{CAS(ME)\textsuperscript{2}} & Part A & 22    & 87    & 30    & S & Yes   & 4 & \multirow{2}{*}{macro/micro} & \multirow{2}{*}{onset,offset,apex} \\
          & Part B & 22    & 57    & 30    & S & Yes   & 4 & \\\hline
    SAMM  &       & 32    & 159   & 200   & S & Yes   & 7$^\ddagger$ & macro/micro & onset,offset,apex \\ \hline
    MEVIEW  &  & 16 & 31 & 25 & S & Yes  & 5$^\S$ & macro/micro & onset,offset \\ \hline
    \bottomrule
    \end{tabular}}
    \newline \raggedright \footnotesize 
    $^\ast$P/S : Posed/Spontaneous \\
    $^\dagger$Not all samples contain micro-expressions and only a total of 183 occurrences of ``micro-tensions'' were annotated. No emotion classes were available.\\
    $^\ddagger$7 objective classes are also provided \citep{davison2017objective}.\\
    $^\S$Set of emotions are atypical (contempt, surprise, fear, anger, happy), likely in the context of environment. Some sample clips involve person speaking, or only have AUs marked with no emotions observed.
  \label{tab:db}%
\end{table}%

\begin{table}[htbp] \footnotesize
  \centering
  \caption{A survey of pre-processing techniques applied in facial micro-expression spotting.}
    \resizebox{\textwidth}{!}{%
    \begin{tabular}{l|c|c|c|c|c}
    \hline \hline
   \multirow{2}[2]{*}{\textbf{Work}} & \textbf{Landmark 
} &\textbf{Landmark 
} & \textbf{Face 
} & \multirow{2}[2]{*}{\textbf{Masking}} & \textbf{Face} \\
 
     & \textbf{Detection} & \textbf{Tracking} & \textbf{Registration} &  & \textbf{Regions} \\ \hline  \hline
   \cite{polikovsky2009facial}  & Manual &  - & - &  - & 12 ROIs \\ \hline
  \cite{shreve2009towards}   &  - &  - &  - & - & 3 ROIs  \\ \hline
   \cite{wu2011}  & - & - &  - &  - & Whole face \\ \hline
    \multirow{2}[2]{*}{\cite{shreve2011macro}}& \multirow{2}[2]{*}{ -} & \multirow{2}[2]{*}{ -} & \multirow{2}[2]{*}{Face alignment} & Eyes, nose & \multirow{2}[2]{*}{8 ROIs} \\
     &  &  &  &  and mouth &  \\ \hline
   \cite{polikovsky2013facial} & Manual & APF & - &  - &12 ROIs \\ \hline
   \multirow{2}[2]{*}{\cite{shreve2014automatic}} & \multirow{2}[2]{*}{SCMS} & \multirow{2}[2]{*}{ -} & \multirow{2}[2]{*}{ -} &Eyes and  & \multirow{2}[2]{*}{4 Parts} \\
     &  &  &  & mouth &  \\ \hline
    \cite{moilanen2014spotting} & Manual & KLT & Face alignment &  - & 6x6 blocks \\ \hline
   \cite{davison2015micro}  & Face++  & - & Affine transform &  - & 5x5 blocks \\ \hline
      \cite{patel2015spatiotemporal}  & DRMF  & OF & - &  - & 49 ROIs \\ \hline
      \cite{liong2015automatic}   & DRMF & - &  &  - & 3 ROIs  \\ \hline
   \multirow{2}[2]{*}{\cite{wang2016main} } & \multirow{2}[2]{*}{DRMF} & \multirow{2}[2]{*}{ -} & Non-reflective & \multirow{2}[2]{*}{ -} & \multirow{2}[2]{*}{6x6 blocks} \\
   &  &  & similarity transformation &  &  \\ \hline
  \cite{liong2016automatic}   & DRMF & - &  - & Eyes & 3 ROIs  \\ \hline
   \cite{xia2016spontaneous}    & ASM & - & Procrutes analysis & - & Whole face \\ \hline
    \cite{liong2016less} & DRMF  & - & - &  - & 3 ROIs  \\ \hline
\cite{davison2016samm}    & Face++ &  - & Affine transform  &  - & 4x4, 5x5 blocks \\ \hline
    \multirow{2}[1]{*}{\cite{davison2016objective}} & \multirow{2}[1]{*}{Face++ } & \multirow{2}[1]{*}{ -} & 2D-DFT and 
 & \multirow{2}[1]{*}{Binary masking} & \multirow{2}[1]{*}{26 ROIs} \\
     &  &  & Piecewise affine warping &  &  \\ \hline
     \cite{yan2017measuring}  & CLM & - & - &  - & 16 ROIs \\ \hline
 \cite{li2017towards}  & Manual & KLT & - &  - & 6x6 blocks \\ \hline
   \cite{ma2017region}  & CLNF & KLT & - &  - & 5 ROIs \\ 
   & from OpenFace  &  &  &  &  \\ \hline
  \cite{qu2017cas}  & ASM & - & LWM &  - & Various block sizes \\ \hline
  \cite{duque2018micro}  & AAM & KLT & - &  - & 5 ROIs \\ \hline
   \hline
    \end{tabular}}%
  \label{tab:tab1}%
\end{table}%

\begin{table}[htbp]
  \centering
  \caption{Facial micro-expression (or micro-movement) spotting works in literature}
  \resizebox{\textwidth}{!}{%
  %\scalebox{0.62}{
    \begin{tabular}{l|c|c|c|c|c}
     \hline \hline
    \textbf{Work} & \textbf{Feature} & \textbf{Feature} & \textbf{Movement (M) /} & \textbf{Spotting Technique} & \textbf{Database} \\
    & & \textbf{Analysis} & \textbf{Apex (A)}& & \\
     \midrule \hline
   \multirow{2}[0]{*}{\cite{polikovsky2009facial}} & \multirow{2}[0]{*}{3D gradient histogram} & \multirow{2}[0]{*}{ -} &  & \multirow{2}[0]{*}{k mean cluster} & High-speed ME database  \\
    &       &       &    &   & (not available) \\ \hline
    \cite{shreve2009towards}        & Optical strain &  -    & M & Threshold technique & USF  \\ \hline
       \cite{wu2011}     & Gabor features &  - & M  & GentleSVM & METT (48 videos) \\ \hline
   \multirow{3}[0]{*}{\cite{shreve2011macro}} & \multirow{3}[0]{*}{Optical strain} & \multirow{3}[0]{*}{ -} &  & \multirow{3}[0]{*}{Threshold technique} & USF-HD \\
   	 &       &       &  M &    & Canal-9 (not available) \\ 
     &       &       &   &    & Found videos (not available) \\ \hline
   \multirow{2}[0]{*}{\cite{polikovsky2013facial}} & \multirow{2}[0]{*}{3D gradient histogram} & \multirow{2}[0]{*}{ -} & & \multirow{2}[0]{*}{k mean cluster} & High-speed ME database  \\
    &       &       &   &    & (not available) \\ \hline
    \multirow{2}[0]{*}{\cite{shreve2014automatic}} & \multirow{2}[0]{*}{Optical strain} & \multirow{2}[0]{*}{ -} & \multirow{2}[0]{*}{M} & \multirow{2}[0]{*}{Threshold technique} & USF \\
     &       &       &  &     & SMIC \\ \hline
    \multirow{3}[0]{*}{\cite{moilanen2014spotting}} & \multirow{3}[0]{*}{LBP} & \multirow{2}[0]{*}{ \checkmark} & &  \multirow{3}[0]{*}{Threshold technique} & CASME-A \\
   	 &       &       &  M  &   & CASME-B \\ 
     &       &       &    &   & SMIC-VIS-E \\ \hline
    \cite{davison2015micro} & HOG & \checkmark & M & Threshold technique & SAMM \\ \hline
    \cite{patel2015spatiotemporal} & Spatio-temporal integration & - & \multirow{2}[0]{*}{M} & Threshold technique & SMIC-VIS-E \\
     &  of OF vectors     &       &    &   & \\ \hline
      \multirow{3}[0]{*}{\cite{liong2015automatic}} & LBP correlation & \multirow{3}[0]{*}{-} & &  \multirow{3}[0]{*}{Binary search} & \multirow{3}[0]{*}{CASME II} \\ 
     &  CLM  &       &  A  &   &  \\ 
     &  Optical strain &       &   &    &  \\ \hline
     \cite{wang2016main} & MDMD & \checkmark & M & Threshold technique & CAS(ME)\textsuperscript{2} \\ \hline
     \multirow{2}[0]{*}{\cite{xia2016spontaneous}} & {Geometrical motion} & \multirow{2}[0]{*}{ -} & \multirow{2}[0]{*}{M} &  \multirow{2}[0]{*}{Random walk model} & CASME \\
     &  deformation  &       &   &    & SMIC \\ \hline
    \cite{liong2016less}    & LBP correlation & -  &  A & Binary search & CASME II \\ \hline
    \multirow{2}[0]{*}{\cite{liong2016automatic}} & LBP correlation & \multirow{2}[0]{*}{-} & \multirow{2}[0]{*}{A} & \multirow{2}[0]{*}{Binary search} & \multirow{2}[0]{*}{CASME II } \\
    & Optical strain &       &   &    &  \\ \hline
        \cite{davison2016samm}   & HOG   & \checkmark & M & Threshold technique & SAMM \\ \hline
   \multirow{3}[0]{*}{\cite{davison2016objective}} & 3D HOG  & \multirow{3}[0]{*}{\checkmark} & & \multirow{3}[0]{*}{Threshold technique} & \multirow{3}[0]{*}{SAMM}  \\
     &  LBP  &       &  M &    & \multirow{3}[0]{*}{CASME II} \\
     & OF  &       &    &   &  \\ \hline
    \multirow{4}[0]{*}{\cite{li2017towards}} & \multirow{4}[0]{*}{ HOOF} & \multirow{4}[0]{*}{\checkmark} &  &  \multirow{4}[0]{*}{Threshold technique} & CASME II \\ 
     & \multirow{4}[2]{*}{ LBP} &  & M &  & SMIC-E-HS  \\ 
     &   &       &    &   & SMIC-E-VIS \\ 
      &   &       &    &   & SMIC-E-NIR \\ \hline
    \multirow{3}[0]{*}{\cite{yan2017measuring}} & LBP correlation & \multirow{3}[0]{*}{-} & &  \multirow{3}[0]{*}{Peak detection} & \multirow{3}[0]{*}{CASME II } \\
    &  CLM  &       &  A  &   &  \\
     & HOOF  &       &    &   &  \\ \hline
  \multirow{2}[0]{*}{\cite{ma2017region}}  & \multirow{2}[0]{*}{RHOOF}   & - & \multirow{2}[0]{*}{A} & \multirow{2}[0]{*}{Threshold technique} & CASME \\ 
  &   &       &    &   & CASME II  \\ \hline
\cite{qu2017cas}  & LBP   & \checkmark & M & Threshold technique & CAS(ME)\textsuperscript{2} \\ \hline
\multirow{2}[0]{*}{\cite{duque2018micro}}  & \multirow{2}[0]{*}{Riesz Pyramid}   & \multirow{2}[0]{*}{\checkmark} & \multirow{2}[0]{*}{M} & \multirow{2}[0]{*}{Threshold technique} & SMIC-E-HS \\
  &   &       &    &   & CASME II  \\ \hline
    \hline
    \end{tabular}}
  \label{tab:spottingTable}%
\end{table}%

\begin{table}
  \centering
  \caption{Benchmarking facial micro-expression recognition works in literature}
  \resizebox{\textwidth}{!}{%
  %\scalebox{0.72}{
    \begin{tabular}{l|c|c|c|c|c|c|c}
    \hline \hline
    \multirow{2}[0]{*}{\textbf{Papers}} & \multirow{2}[0]{*}{\textbf{Pre-processing}} & \multirow{2}[0]{*}{\textbf{Features}} &  \multirow{2}[0]{*}{\textbf{Classifier}} & \multicolumn{2}{c}{\textbf{Accuracy (\%)}} \vline & \multicolumn{2}{c}{\textbf{F1-score (\%)}} \\
  
    &  &       &     & \textbf{CASME II} & \textbf{SMIC} & \textbf{CASME II} & \textbf{SMIC} \\ 
   \hline
   \multicolumn{8}{l}{\rule{0pt}{1em}\textbf{LOSO}\rule[-0.5em]{0pt}{0pt}} \\
   \hline
   \cite{li2013spontaneous} & - & LBP-TOP & SVM   & -     & 48.78 & -     & - \\ \hline
    \cite{liong2016spontaneous} & - & OSF + OS weighted LBP-TOP & SVM   & -     & 52.44 & -     & - \\ \hline
    \cite{liong2014optical} & - & OS  & SVM   & -     & 53.56 & -     & - \\ \hline
    \cite{liong2014subtle} & - & OS weighted LBP-TOP   & SVM   & 42.00     & 53.66 & 0.38     & 0.54 \\ \hline
    \cite{le2014spontaneous} & - & STM & Adaboost & 43.78 & 44.34 & 0.3337 & 0.4731  \\ \hline 
    \cite{wang2015efficient} & - & LBP-MOP  & SVM   & 44.13 & 50.61 & -     & - \\ \hline
    \cite{xu2016microexpression} & - & Facial Dynamics Map  & SVM   & 45.93 & 54.88 & 0.4053 & 0.538 \\ \hline
    \cite{oh2016intrinsic} & - & Monogenic + LBP-TOP & SVM   & -     & -     & 0.41  & 0.44 \\ 
  %  &  & + LBP-TOP &   &      & -     &  & \\ 
  \hline
    \cite{oh2015monogenic} & - & Riesz wavelet + LBP-TOP  & SVM   & -     & -     & 0.43 & - \\ \hline
    \cite{liong2017hybrid} & ROIs & LBP-TOP & SVM & 46.00 & 54.00 & 0.32  & 0.52 \\ \hline
    \cite{wang2014lbp} & - & LBP-SIP & SVM & 46.56 & 44.51 & 0.448  & 0.4492 \\ \hline
    \cite{le2016eulerian} & A-EMM & LBP-TOP & SVM   & -     & -     & 0.51  & - \\ \hline
	\cite{le2016sparsity} & DMDSP & LBP-TOP & SVM   & 49.00    & 58.00    & 0.51  & 0.60 \\ \hline
    \cite{park2015subtle} & Adaptive MM & LBP-TOP & SVM  & 51.91    & -    & -  & - \\ \hline
    \cite{happy2017fuzzy} & - & HFOFO  & SVM   & 56.64 & 51.83  & 0.5248  & 0.5243 \\ \hline 
    \cite{liong2016less} & - & Bi-WOOF & SVM   & -     & -     & 0.56  & 0.53 \\ \hline
    \cite{huang2016spontaneous} & - & STCLQP  & SVM   & 58.39 & 64.02 & 0.5836 & 0.6381 \\ \hline
    \cite{huang2015facial} & - & STLBP-IP  & SVM   & 59.51 & 57.93 & 0.57$^{*}$ & 0.58$^{*}$ \\ \hline
    \cite{liong2016automatic} & - & Bi-WOOF (apex frame) & SVM   & -     & -     & 0.61  & 0.62 \\  \hline
    \cite{he2017multi} & - & MMFL & SVM   & 59.81 & 63.15 & -     & - \\ \hline
    \cite{kim2016micro} & - & CNN + LSTM & Softmax & 60.98 & - & - & - \\ \hline
    \cite{liong2017micro} & - & Bi-WOOF + Phase & SVM & 62.55 & 68.29 & 0.65 & 0.67 \\ \hline
    \cite{zheng2016relaxed} & - & LBP-TOP  & RK-SVD & 63.25 &       & -     & - \\ \hline
    \cite{zong2018learning} & - & Hierarchical STLBP-IP & KGSL   & 63.83 & 60.78     & 0.6110     & 0.6126 \\ \hline
    \cite{huang2017spontaneous} & TIM & STRBP & SVM   & 64.37   & 60.98     & -  & - \\  \hline
    \cite{xiaohua2017discriminative} & - & Discriminative STLBP-IP & SVM   & 64.78 & 63.41 & -     & - \\ \hline
    \cite{allaertconsistent} & - & OF Maps & SVM   & 65.35 & -     & -     & - \\ \hline
    \cite{li2017towards} & TIM+EVM & HIGO  & SVM   & 67.21 & 68.29 & -     & - \\ \hline
    \cite{zheng2017micro} $^\dagger$$^\ddagger$  & - & 2DSGR  & SRC   &  -     & 71.19 & -     & - \\ \hline
	\cite{liu2016main} $^\dagger$ & - & MDMO   & SVM   & 67.37 & 80.00    & -     & - \\ \hline
    %\cite{lu2014delaunay} $^\dagger$ & TIM & DCTM  & SVM (SMO) & 72.06 & 82.86 & -     & - \\ \hline
    \cite{davison2017objective} $^\ddagger$ & - & HOOF & SVM & 76.60 & - & 0.55  & - \\  \hline
    \multicolumn{8}{l}{\rule{0pt}{1em}\textbf{LOVO}\rule[-0.5em]{0pt}{0pt}} \\
    \hline
    \cite{wang2015micro} $^\dagger$$^\ddagger$ & TIM & LBP-TOP on TICS & SVM & 62.30  & -     &   -    & -  \\ 
	%& & feature learning  &   & & &      &  \\ 
    \hline
   %\cite{liong2016spontaneous} & - & OS + weighted LBP-TOP & SVM   & 63.16 & -     & -     & - \\ \hline
    \cite{yan2014casme} & - & LBP-TOP & SVM   & 63.41 & -     & -     & - \\ \hline
    \cite{wang2014micro} & TIM & DLSTD & SVM   & 63.41 & 68.29 & -     & - \\ \hline
    \cite{happy2017fuzzy} & - & HFOFO  & SVM   & 64.06 & 56.10  & 0.6025  & 0.5536 \\ \hline 
   \cite{liong2014subtle} & - & OS weighted LBP-TOP  & SVM   & 65.59 & -     & -     & - \\ \hline
    \cite{wang2015efficient} & - & LBP-MOP  & SVM   & 66.80  & 60.98 & -     & - \\ \hline
    \cite{wang2014lbp} & - & LBP-SIP  & SVM   & 67.21 & -     & -     & - \\ \hline
   \cite{ping2016micro} &       & LBP-TOP & GSLSR & 67.89 & 70.12 & -     & - \\ \hline
   \cite{park2015subtle} & Adaptive MM & LBP-TOP & SVM  & 69.63    & -    & -  & - \\ \hline
    \cite{wang2016effective} & EVM & LBP-TOP  & SVM   & 75.30  & -     & -     & - \\ \hline
    \cite{li2017towards} & TIM+EVM & HIGO  & SVM   & 78.14 & 75.00    & -     & - \\ \hline 
    \hline 
    \multicolumn{8}{l}{\rule{0pt}{1em}\textbf{OTHER PROTOCOLS} \rule[-0.5em]{0pt}{0pt}} \\
    \hline
    \cite{zhang2017micro} & \multirow{2}{*}{-} & \multirow{2}{*}{LBP-TOP and HOOF} & \multirow{2}{*}{RF} & \multirow{2}{*}{62.5}  & \multirow{2}{*}{-}   & -\multirow{2}{*}{-}  & \multirow{2}{*}{-} \\
    \quad \textit{Evenly Distributed} & & & & & & & \\ \hline
    \cite{jia2017macro} &   \multirow{2}{*}{-}    & \multirow{2}{*}{SVD+ LBP/LBP-TOP}  & \multirow{2}{*}{KNN}   & \multirow{2}{*}{65.5} & \multirow{2}{*}{-}  & \multirow{2}{*}{-}  & \multirow{2}{*}{-} \\ 
    \quad \textit{Random Test (20 times)} & & & & & & & \\ \hline
    \cite{peng2017dual} $^\S$$^\ddagger$ &   \multirow{2}{*}{-}    & \multirow{2}{*}{DTSCNN}  & \multirow{2}{*}{SVM}   & \multirow{2}{*}{66.67} & \multirow{2}{*}{-}  & \multirow{2}{*}{-}  & \multirow{2}{*}{-} \\ 
    \quad \textit{3-fold cross-validation} & & & & & & & \\ \hline
    \cite{adegun2016automatic} $^\dagger$ & \multirow{2}{*}{-}       & \multirow{2}{*}{LBP-TOP} & \multirow{2}{*}{ELM}   & \multirow{2}{*}{96.12} & \multirow{2}{*}{-}     & \multirow{2}{*}{-}     & \multirow{2}{*}{-} \\
    \quad \textit{5-fold cross-validation} & & & & & & & \\ \hline
    \bottomrule
  \multicolumn{8}{l}{\small $\dagger$\hspace{1.5em} Not all the samples in the dataset were used in the experiments.} \\ 
  \multicolumn{8}{l}{\small $\ddagger$\hspace{1.5em} Different number of emotion classes were used in the experiments.} \\
  \multicolumn{8}{l}{\small $\S$\hspace{1.5em} Combined CASME I/II database was used.} \\
	%\multicolumn{8}{l}{\small $\star$\hspace{1.5em} Different evaluation protocol was applied compared to that used by majority of works listed in the table.} \\
  \multicolumn{8}{l}{\small $*$\hspace{1.5em} Result not reported in paper, but computed from confusion table provided.} \\
%  \multicolumn{8}{l}{\small $\dagger$ $\star$ \hspace{0.4em} Not all the samples in the dataset were used and different evaluation protocol was applied compared to the ones used in majority of the work in the table.} \\
    %\multicolumn{8}{l}{\small $\upsilon$\hspace{1.5em} Different objective class labels.} \\
    \end{tabular}}
  \label{tab:recogTable}
\end{table}%

\clearpage

\section*{Figures} \label{sec:fig}
\addcontentsline{toc}{section}{Figures}

\begin{figure}[!ht]
\centering
\includegraphics[scale=0.97]{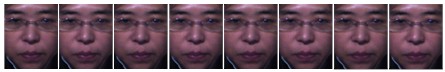}
\caption{Sample frames from a 'Surprise' sequence (Subject 1) in SMIC. Images reproduced from the database with permission from \citep{li2013spontaneous}.}
\label{fig:smic_sur}
\end{figure} 

\begin{figure}[!ht]
\centering
\includegraphics[scale=0.65]{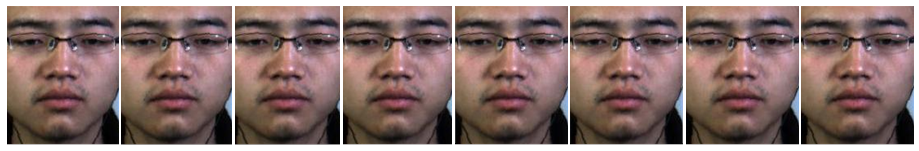}
\caption{Sample frames from a 'Happiness' sequence (Subject 6) in CASME II. Images reproduced from the database with permission from \citep{yan2014casme}.}
\label{fig:casme_hap}
\end{figure}

\begin{figure}[!ht]
\centering
\includegraphics[scale=0.98]{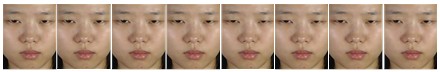}
\caption{Sample frames from a 'Disgust' sequence (Subject 15) in CAS(ME)\textsuperscript{2}. Images reproduced from the database (\textcopyright Xiaolan Fu) with permission from \citep{qu2017cas}.}
\label{fig:casme2_disgust}
\end{figure}

\begin{figure}[!t]
\centering
\includegraphics[scale=0.72]{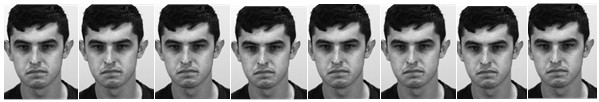}
\caption{Sample frames from a sequence (Subject 6) in SAMM that contains micro-movements. Images reproduced from the database with permission from \cite{davison2016samm}.}
\label{fig:samm}
\end{figure}

\begin{figure}[!t]
\centering
\includegraphics[width=150mm]{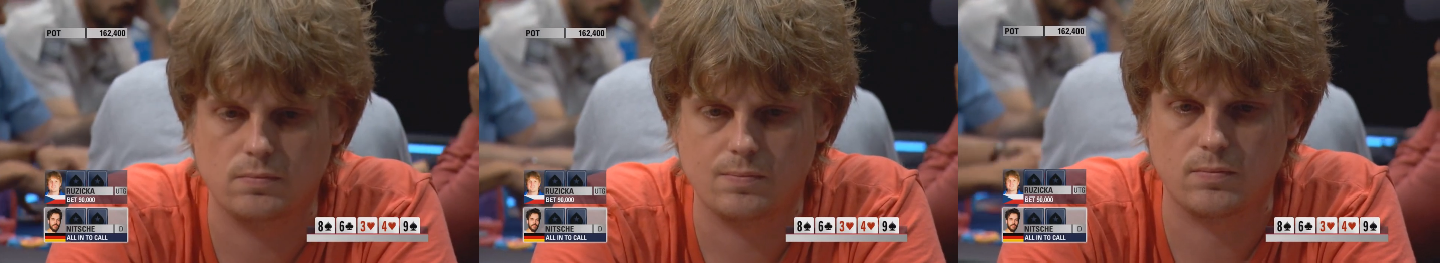}
\caption{Sample frames from a 'Contempt' sequence in MEVIEW that contains micro-movements marked with AU L12. Images reproduced from the database \citep{meview} under Fair Use.}
\label{fig:meview}
\end{figure}

\begin{figure}[!t]
\centering
\includegraphics[width=150mm]{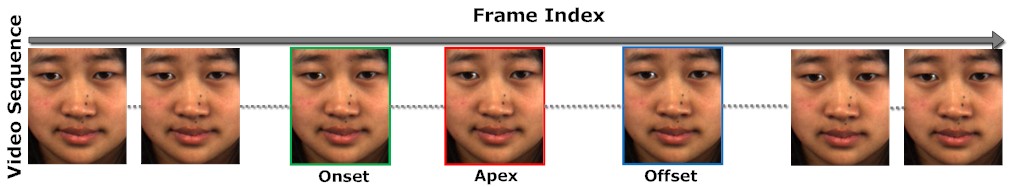}
\caption{A video sequence depicting the order of which onset, apex and offset frames occur. Sample frames are from a "Happiness" sequence (Subject 2) in CASME II. Images reproduced from the database with permission from \cite{yan2014casme}.}
\label{fig:spotting}
\end{figure}

\clearpage

% \section*{Appendix A. Placeholder} \label{sec:appendixa}
% \addcontentsline{toc}{section}{Appendix A}

\end{document}